\documentclass[journal]{IEEEtran}

\usepackage{microtype}
\usepackage{graphicx}
\usepackage{subcaption}
\usepackage{booktabs}
\usepackage{multirow}
\usepackage[table]{xcolor}
\usepackage{color}
\usepackage[percent]{overpic}
\usepackage{pifont}
\usepackage{float}
\usepackage{siunitx}
\usepackage{placeins}
\usepackage{amsmath}
\usepackage{amssymb}
\usepackage{mathtools}
\usepackage{amsthm}
\usepackage[numbers,sort&compress]{natbib}
\usepackage[colorlinks,
            linkcolor=blue,
            anchorcolor=blue,
            citecolor=blue]{hyperref}
\usepackage[capitalize,noabbrev]{cleveref}

\graphicspath{{images/}}

\theoremstyle{plain}

\theoremstyle{definition}

\theoremstyle{remark}

\crefname{section}{Sec.}{Secs.}
\Crefname{section}{Section}{Sections}
\Crefname{table}{Table}{Tables}
\crefname{figure}{Fig.}{Figs.}

\begin{document}

\title{Beyond Proposals: Region-Consistent Open-Set Object Detection}

\author{Jiehui Tang,
        Yunlong Yu$^\dagger$,
        Zhao Wang,
        and~Jungong Han
        

\thanks{Yunlong Yu$^\dagger$ is the corresponding author.}
\thanks{Jiehui Tang and Yunlong Yu are with the College of Information Science and Electronic Engineering, Zhejiang University, Hangzhou, 310027 China (e-mail: \{tangjiehui, yuyunlong\}@zju.edu.cn).}
\thanks{Zhao Wang is with Ningbo Innovation Center, Zhejiang University, Ningbo, China (e-mail: zhao\_wang@zju.edu.cn). }
\thanks{Jungong Han is with the Department of Automation, Tsinghua University, Beijing 100084 (e-mail: jungonghan77@gmail.com).}

}


\maketitle

\begin{abstract}

Open-set object detection (OSOD) suffers from a fundamental limitation under dense proposal mechanisms: highly overlapping proposals associated with the same object are often assigned contradictory known/unknown predictions. Existing methods perform open-set reasoning independently for each proposal, while ignoring the shared semantics among correlated proposals, resulting in fragmented detections and unstable open-set decisions. In this paper, we reformulate OSOD as an object-centric semantic inference problem, where correlated proposals jointly support a coherent region-level semantic decision. Based on this perspective, we propose OCSI-Net, a unified framework that combines object-centric semantic consistency with geometry-aware uncertainty stabilization. Specifically, OCSI-Net aggregates correlated proposals into structured region clusters and infers region-level semantics from aggregated semantic evidence, thereby enforcing coherent open-set predictions within the same object region. Furthermore, a geometry-aware uncertainty mechanism is introduced to stabilize optimization near open-set transition boundaries. Extensive experiments on multiple benchmarks demonstrate that OCSI-Net consistently achieves state-of-the-art performance, improving the harmonic mean of known and unknown detection by an average of 6.83\% over the second-best method.

\end{abstract}


\IEEEpeerreviewmaketitle

\section{Introduction}
\label{sec:introduction}

\IEEEPARstart{O}{bject} detection has achieved remarkable success under the closed-set assumption, where training and testing share identical category sets. However, real-world environments are inherently open, and unseen object categories frequently appear during deployment. Open-set object detection (OSOD)~\citep{hosoya25rethinkingosod,Han_2022_CVPR,Park_2025_ICCV_SFUOD,Yuan_2025_ICCV_ASGS} addresses this challenge by requiring detectors to simultaneously recognize known objects while identifying objects belonging to unknown categories.

\begin{figure}[!t]
    \centering
    \begin{overpic}[width=1\linewidth]{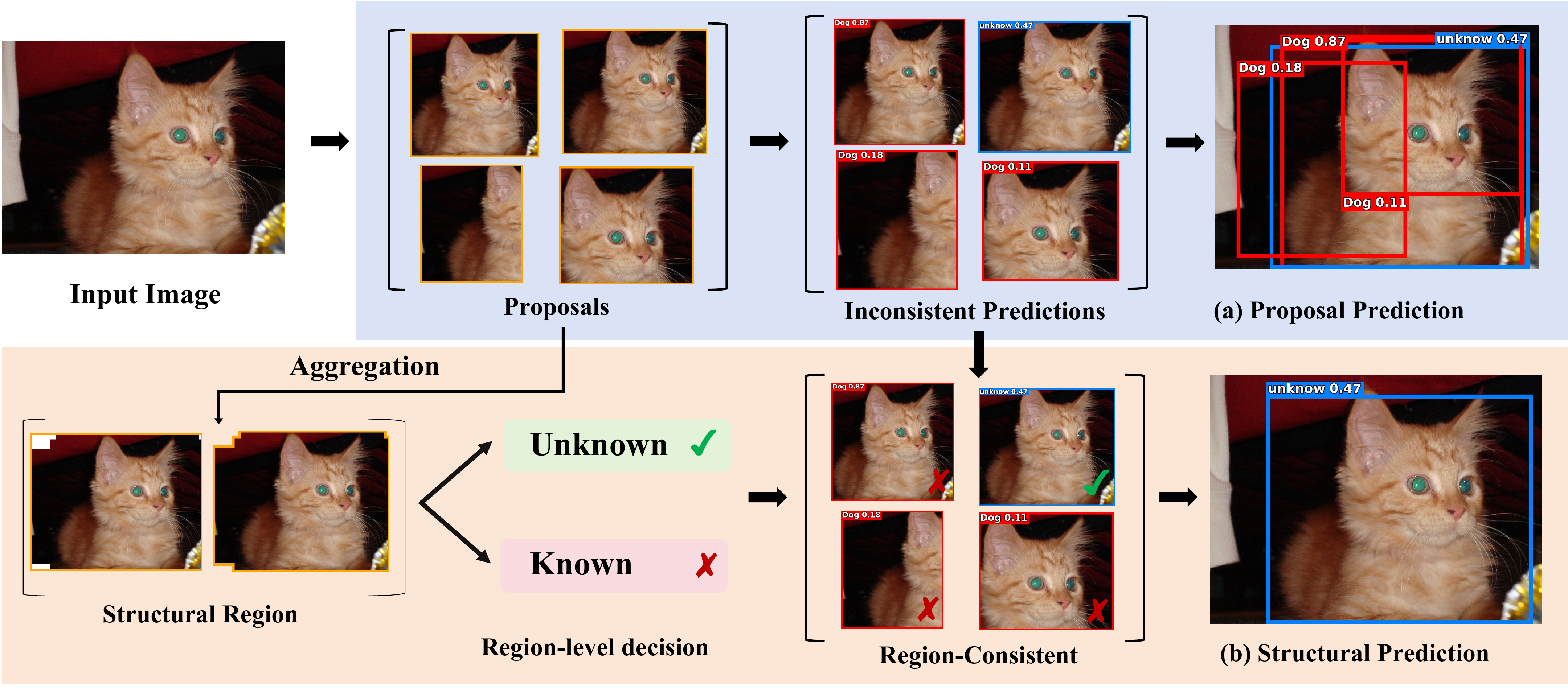}
    \end{overpic}
    \caption{(a) Proposal prediction: independent decisions over dense RoIs lead to conflicting known (red) and unknown (blue) predictions within the same object region. (b) Structural prediction: aggregating proposals into region clusters resolves semantic conflicts and yields coherent detections.}
    \label{fig:samples}
\end{figure}

Existing OSOD methods mainly extend modern two-stage detectors by introducing unknown-aware scoring functions, feature separation objectives, uncertainty estimation strategies, or contrastive representation learning mechanisms~\citep{Han_2022_CVPR,Wu_2025_CVPR_WFS,Yuan_2025_ICCV_ASGS}. Despite their effectiveness, most approaches still inherit the dense proposal mechanisms of contemporary detectors~\citep{ren2015faster,qiao2021detectors}, where a large number of highly overlapping proposals are generated around each object instance. Open-set reasoning is then performed independently for each proposal.

While such a paradigm is effective for conventional closed-set classification, it introduces a fundamental yet overlooked problem in open-set scenarios: \emph{object-level semantic consistency among dense proposals is severely broken}. As illustrated in Fig.~\ref{fig:samples}(a), a single object is usually covered by many geometrically aligned and visually similar proposals, which should naturally share the same semantic identity. However, the existing OSOD methods frequently produce contradictory known/unknown predictions within the same object region, where some proposals are confidently classified as known while neighboring proposals are simultaneously predicted as unknown. Such inconsistencies lead to fragmented detections and unstable open-set predictions that cannot be effectively resolved by conventional post-processing strategies such as non-maximum suppression.

We argue that this phenomenon is not an incidental failure case, but rather a structural limitation caused by the mismatch between \emph{dense proposal sampling} and \emph{proposal-level independent open-set decision making}. From a probabilistic perspective, proposals associated with the same object should instead be regarded as multiple noisy observations of a shared latent semantic state. However, existing methods ignore this dependency and treat proposals as mutually independent decision units. Consequently, proposal-level predictions become highly unstable near semantic transition boundaries. In particular, small geometric perturbations around object boundaries may trigger abrupt transitions between known and unknown predictions, thereby amplifying semantic inconsistency within the same object region.

This observation reveals two critical limitations of existing OSOD methods. First, although multiple proposals may correspond to the same underlying object, their shared semantic dependency is largely ignored during open-set reasoning. Existing approaches mainly focus on proposal-level feature separation or uncertainty estimation, while overlooking object-centric semantic consistency among correlated proposals. Second, semantic uncertainty in OSOD is coupled with proposal geometry. Unlike closed-set detection, OSOD must additionally model the semantic transition boundary between known and unknown regions. As a result, proposals near geometric transition regions often exhibit severe semantic ambiguity and prediction instability, making open-set optimization sensitive to localization perturbations.

Motivated by these observations, we reformulate OSOD as an \emph{object-centric semantic inference problem}, where geometrically correlated proposals are jointly required to support a coherent region-level semantic decision. Under this formulation, two key challenges naturally arise: 
(i) how to infer stable object-level semantics from noisy and potentially conflicting proposal observations; and 
(ii) how to stabilize open-set optimization near geometric transition boundaries where semantic uncertainty becomes highly sensitive to spatial perturbations.

To address these challenges, we propose Object-Centric Semantic Inference Network (OCSI-Net), a unified object-centric open-set object detection framework that jointly models Object-Centric semantic consistency and geometry-aware uncertainty stabilization. Instead of treating proposals as isolated decision units, OCSI-Net models proposals within the same object region as correlated observations that share a latent semantic state.

Specifically, OCSI-Net consists of two coupled components. First, we introduce an Object-Centric Semantic Consistency (OCSC) module, which performs object-centric semantic inference by aggregating correlated proposals into geometry-consistent region clusters and inferring unified region-level semantics from aggregated semantic evidence. By promoting open-set reasoning from proposal-level independent classification to structured object-centric semantic inference, OCSC suppresses contradictory known/unknown predictions within the same object region and improves semantic coherence. To further stabilize object-centric semantic inference, we introduce a Geometry-Aware Boundary Uncertainty (GABU) mechanism, which explicitly models geometry-induced semantic uncertainty near open-set transition boundaries. Instead of uniformly optimizing all proposals, GABU dynamically emphasizes geometrically ambiguous proposals that dominate boundary uncertainty, thereby stabilizing unknown-probability optimization and learning smoother and more robust decision boundaries.

In summary, our contributions are threefold:
\begin{itemize}

\item We reformulate OSOD as an object-centric semantic inference problem and propose OCSI-Net, a unified framework that jointly addresses semantic inconsistency and geometry-induced semantic instability under dense proposal sampling.

\item We develop Object-Centric Semantic Consistency (OCSC) and Geometry-Aware Boundary Uncertainty (GABU) for object-level semantic reconciliation and uncertainty-aware boundary stabilization.

\item Extensive experiments on multiple benchmarks demonstrate that OCSI-Net consistently achieves state-of-the-art performance, substantially improving unknown-object detection while preserving strong known-category accuracy. In particular, OCSI-Net improves the harmonic mean of known and unknown detection by an average of \textbf{6.83\%} over the second-best method across multiple benchmarks.

\end{itemize}

The remainder of this paper is organized as follows. Section~\ref{sec:relatedwork} reviews related work. Section~\ref{sec:method} presents the proposed method. Section~\ref{sec:experiment} provides extensive experiments and analyses. Section~\ref{sec:limitations} discusses limitations and further analysis. Finally, Section~\ref{sec:conclusion} concludes this paper.

\section{Related Works} 
\label{sec:relatedwork}

\noindent\textbf{Open-set Object Detection.}
Open-set object detection (OSOD) studies detectors deployed in open environments where test images may contain objects outside the training label space; such unknown instances can be assigned high confidence as known categories, motivating evaluation beyond closed-set mAP to reveal open-set failure modes \citep{miller18dropout,dhamija20overlooked}.
Recent progress improves unknown handling via unknown-aware scoring and graph-based post-processing \citep{liang23unsniffer}, representation decoupling/orthogonality for incremental learning \citep{Sun_2024_CVPR}, attribute-based reasoning \citep{Yang_2025_CVPR}, and unified formulations bridging open-world and open-vocabulary detection \citep{Xi_2025_CVPR}, as well as explicit open-world objectness modeling \citep{Zhang_2025_CVPR} and behavior-driven unknown decision analysis \citep{Xi_2024_NeurIPS_UMB}.
However, defining unknown as any unseen class renders open-set detection ill-posed with respect to what should be detected \citep{hosoya25rethinkingosod}; accordingly, \citet{hosoya25rethinkingosod} propose OSOD-III, which restricts unknowns to novel sub-classes within user-specified super-classes and treats instances outside the super-class as non-targets.
Even under this more controlled setting, most systems still decide at the single-box level and rely on thresholding/NMS, which may fail to ensure consistent known/unknown decisions among highly overlapping proposals around the same instance \citep{Xi_2024_NeurIPS_UMB}.
In parallel, vision--language pretraining has enabled transferable representations for open-vocabulary/open-set detection \citep{radford21clip,li22glip,zhou22detic,liu24groundingdino,Cheng_2024_CVPR}, yet these methods typically focus on vocabulary expansion and per-box classification, leaving the aggregation of dense, highly-overlapping proposals to conventional thresholding/NMS.


\begin{figure*}[t]
  \centering
  \includegraphics[width=\textwidth]{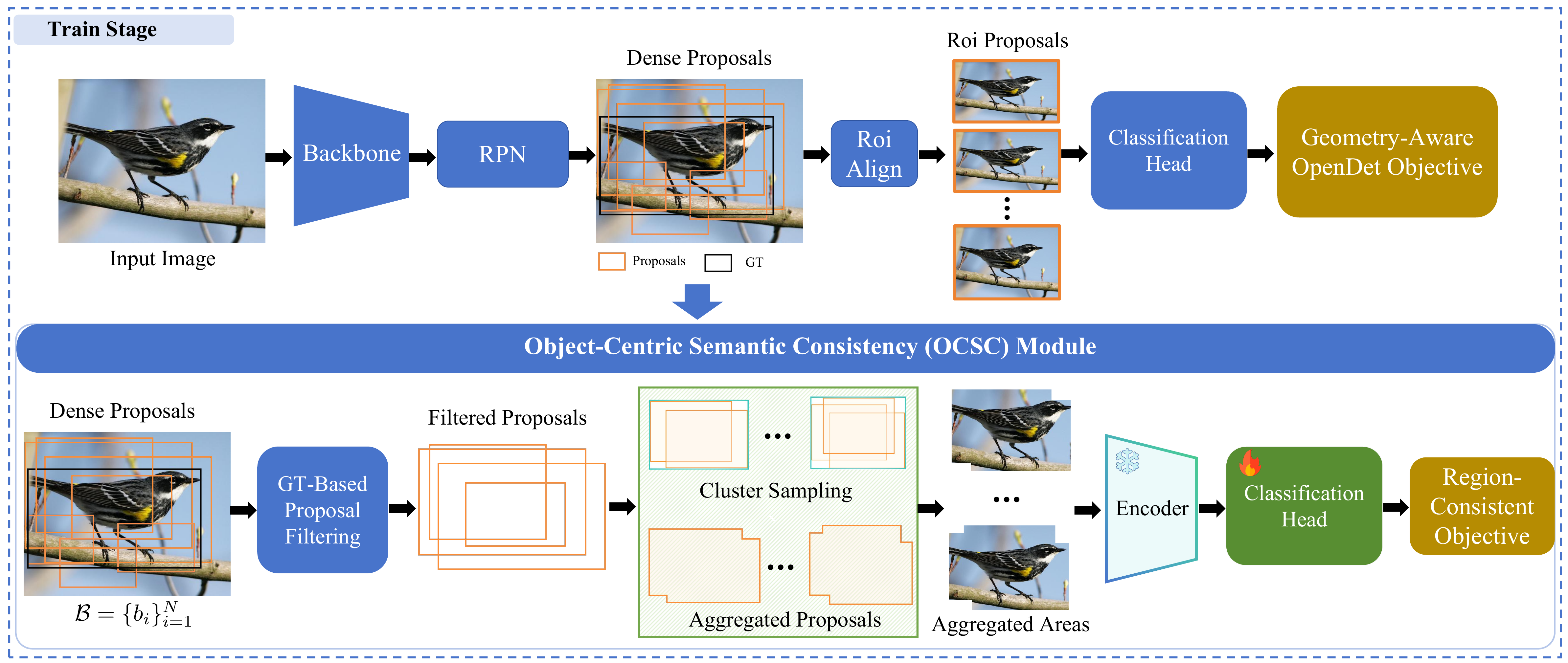}
  \caption{Illustration of the proposed framework during training. The framework first generates dense proposals and subsequently produces proposal-level predictions. The entire detector is optimized under the proposed geometry-aware OpenDet objective, while the Object-Centric Semantic Consistency (OCSC) module is jointly optimized to enforce object-centric semantic coherence among correlated proposals.}
  \label{fig:overview}
\end{figure*}

\noindent\textbf{Proposal Grouping and Post-processing.}
Modern detectors typically produce many redundant and highly overlapping candidate boxes, making post-processing essential to consolidate near-duplicates into final predictions.
Non-maximum suppression (NMS) and its variants remain the dominant choice; Soft-NMS improves robustness under heavy overlaps by decaying scores instead of hard suppression \citep{bodla17softnms}, while clustering-pattern-based variants further target crowded/high-overlap failure modes \citep{Liao_2025_ClusterNMS}.
Beyond hand-crafted rules, learning-based post-processing predicts suppression/merging decisions directly from sets of boxes and scores, reducing reliance on fixed IoU heuristics \citep{hosang17learningnms}.
Clustering-style alternatives provide a more principled replacement of greedy NMS: CP-Cluster performs message passing on a box graph to explicitly group overlapping candidates and refine confidences in parallel \citep{shen22cpcluster}.
In parallel, end-to-end detector designs reduce or remove heuristic post-processing by outputting sparse predictions or improving matching/training dynamics \citep{sun21sparsercnn,carion20detr,zhu21deformabledetr}; real-time end-to-end detectors further highlight the latency/accuracy gains of avoiding NMS \citep{Zhao_2024_CVPR}.
Relatedly, NMS-free one-stage designs aim to eliminate NMS at inference via training mechanisms that enforce ``one positive per instance'' \citep{Wang_2024_NeurIPS_YOLOv10}.
Despite these advances, existing pipelines largely treat single boxes as the basic decision units and focus on de-duplication/sparsification; they rarely impose semantic consistency within a local cluster of highly overlapping proposals, which becomes particularly important when unknown handling is required.

\section{Method}\label{sec:method}

As illustrated in Fig.~\ref{fig:overview}, the proposed method is built upon a standard two-stage open-set detector and reformulates dense proposal reasoning through an Object-Centric Semantic Consistency (OCSC) module, which performs structured object-centric semantic inference over correlated proposals. To further stabilize semantic inference, a Geometry-Aware Boundary Uncertainty (GABU) mechanism is integrated into the training objective, enabling geometry-aware uncertainty stabilization during optimization.



\subsection{Object-Centric Semantic Consistency (OCSC) Module}
\label{sec:ocsc}

OCSC is organized into four stages: structured proposal dependency construction,
cluster-level spatial support construction, region-level semantic evidence extraction,
and semantic consistency inference.


\subsubsection{Structured Proposal Dependency Construction}

During training, the detector first produces a dense proposal set:
\begin{equation}
\mathcal{B}
=
\{b_i\}_{i=1}^{N},
\end{equation}
where \(b_i\) denotes the \(i\)-th dense proposal and \(N\) is the total number of dense proposals in an image. Let \(\mathcal{T}=\{g_j\}_{j=1}^{M}\) denote the set of ground-truth boxes in the image.

For each proposal, we identify its best-matched ground-truth target:
\begin{equation}
j^*(i)
=
\arg\max_j
\mathrm{IoU}(b_i, g_j),
\end{equation}
where \(g_j\) denotes the \(j\)-th ground-truth box, and \(j^*(i)\) is the index of the ground-truth target that has the largest overlap with proposal \(b_i\). This assignment associates each proposal with the object target that provides the strongest geometric support.

As illustrated in \cref{fig:overview}, we first apply the ground-truth object annotations as the target anchors to associate dense proposals with their most relevant object instances, and then filter proposals
that provide reliable geometric support for each target. Specifically, for each
ground-truth target \(g_j\), we construct a target-specific proposal pool \(\mathcal{P}_j\):
\begin{equation}
\mathcal{P}_j
=
\left\{
i
~
\middle|
~
\mathrm{IoU}(b_i,g_j)\ge\tau_{\text{clus}},
~
j^*(i)=j
\right\},
\end{equation}
where \(\mathcal{P}_j\) denotes the target-specific proposal pool associated
with ground-truth target \(g_j\), and \(\tau_{\text{clus}}\) denotes the cluster
construction threshold.

This filtering process imposes two constraints. The overlap
constraint \(\mathrm{IoU}(b_i,g_j)\ge\tau_{\text{clus}}\) ensures that the
selected proposals are geometrically supported by the target,
while the assignment constraint \(j^*(i)=j\) ensures that each proposal is
assigned to its most relevant ground-truth object rather than being ambiguously
shared by nearby targets.

Based on the proposal pool \(\mathcal{P}_j\), we then sample the proposals to obtain the proposal cluster with controlled cluster sizes:
\begin{equation}
\mathcal{G}^{\mathrm{tr}}_k
\subseteq
\mathcal{P}_j,
\qquad
|\mathcal{G}^{\mathrm{tr}}_k|=m_k,
\qquad
k=1,\ldots,K_{\mathrm{tr}} .
\end{equation}
Here, \(\mathcal{G}^{\mathrm{tr}}_k\) denotes the \(k\)-th training proposal
cluster, \(K_{\mathrm{tr}}\) denotes the number of sampled training clusters,
and \(m_k\) denotes the number of proposals sampled from \(\mathcal{P}_j\).
Each sampled training cluster is generated from one target-specific proposal pool \(\mathcal{P}_j\), and the corresponding target index is omitted for notational simplicity. In our implementation, \(m_k\) is
sampled from a small candidate-size set, e.g.,
\(m_k\in\{2,2,3,3,4,5\}\). Thus, \(m_k\) controls the cluster size, rather than indexing the cluster itself.

\begin{figure*}[t]
  \centering
  \includegraphics[width=\textwidth]{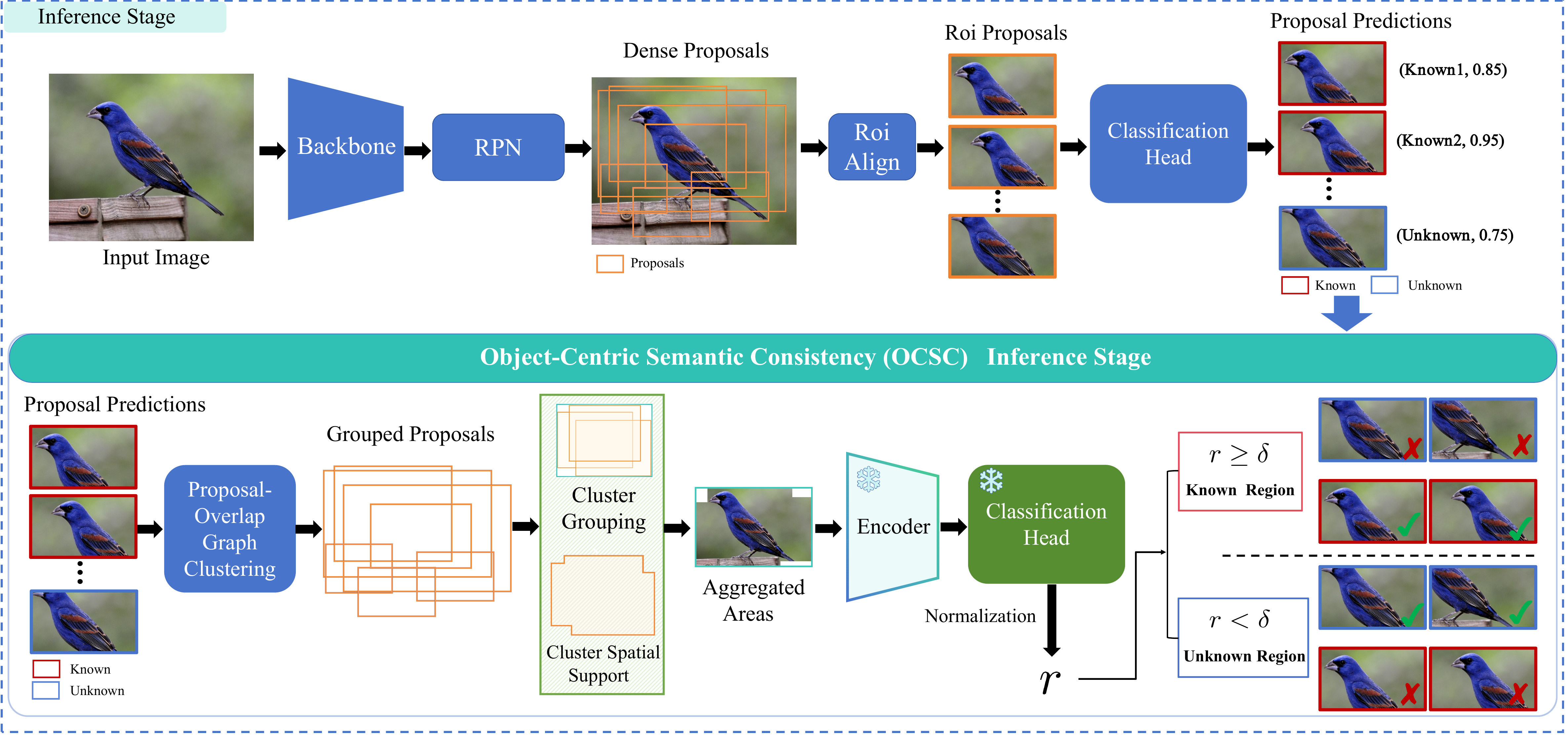}
  \caption{Illustration of the proposed inference framework. Starting from proposal-level predictions, OCSC groups correlated proposals into object-centric region clusters and performs structured semantic inference based on aggregated region evidence. The inferred region semantics are then used to reconcile inconsistent known/unknown responses, producing coherent and region-consistent open-set detections.}

  \label{fig:inference}
\end{figure*}

During inference, object annotations are unavailable, and proposal clusters can no longer be constructed from ground-truth-associated proposal pools. Instead, we build inference clusters directly from proposal predictions according to their
pairwise box-level geometric overlap. As illustrated in \cref{fig:inference}, this process is implemented by grouping proposal predictions according to their mutual spatial overlap without relying on ground-truth targets. Let $\mathcal{D} = \{d_i\}_{i=1}^{n}$ denote the proposal prediction set produced by the detector, where \(d_i\) denotes the \(i\)-th prediction and \(n\) is the number of proposal predictions. Each \(d_i\) contains a predicted box, denoted as \(\mathrm{box}(d_i)\), and its corresponding semantic response.

Specifically, we construct a proposal-overlap graph whose vertices are all proposal predictions in \(\mathcal{D}\). Two proposal predictions are connected if their predicted boxes have sufficient spatial overlap:
\begin{equation}
\mathcal{E}
=
\left\{
(i,a)
~
\middle|
~
\mathrm{IoU}\big(\mathrm{box}(d_i),\mathrm{box}(d_a)\big)
\ge
\tau_{\mathrm{grp}}
\right\},
\end{equation}
where \(\mathcal{E}\) denotes the set of overlap edges, \(a\) indexes another proposal prediction in \(\mathcal{D}\), and \(\tau_{\mathrm{grp}}\) is the inference grouping threshold.

Unlike the target-conditioned filtering strategy used during training, inference-time clustering does not rely on target-specific proposal pools, since object annotations are unavailable. Instead, all proposal predictions are preserved as graph vertices, and proposal grouping is determined solely according to pairwise geometric overlap relations. The inference clusters are built from proposal predictions based on their mutual
box-level geometric overlap:
\begin{equation}
\{\mathcal{G}^{\mathrm{inf}}_s\}_{s=1}^{K_{\mathrm{inf}}}
=
\mathrm{CC}
\left(
\{1,\ldots,n\},\mathcal{E}
\right),
\end{equation}
where \(\mathcal{G}^{\mathrm{inf}}_s\) denotes the \(s\)-th inference cluster, \(K_{\mathrm{inf}}\) denotes the number of inference clusters, and \(\mathrm{CC}(\cdot)\) denotes connected-component grouping over the proposal-overlap graph, returning the set of connected proposal-index components. Therefore, \(K_{\mathrm{inf}}\) is not a predefined hyperparameter, but is determined by the number of connected proposal groups formed in the current image. Multiple objects in one image naturally lead to multiple inference clusters when their proposal predictions are not connected by sufficient spatial overlap. Conversely, a proposal prediction that has no overlap edge with other predictions forms an individual singleton cluster, rather than being discarded.

Both \(\mathcal{G}^{\mathrm{tr}}_k\) and \(\mathcal{G}^{\mathrm{inf}}_s\) are constructed to approximate candidate object regions, where highly overlapping proposals or proposal predictions are treated as correlated observations of the same latent semantic state. Note that the key idea is not simply grouping boxes together, but explicitly modeling semantic dependency among proposals. Unlike conventional NMS, which directly removes redundant proposals during post-processing, the OCSC module first preserves correlated proposals as complementary semantic evidence and performs semantic consolidation at the region level.

\subsubsection{Cluster-Level Spatial Support Construction}

After cluster construction, training and inference clusters follow the same region-level support modeling process. For notational clarity, we use \(\mathcal{G}\) to denote a constructed proposal cluster in the following shared formulation, where \(\mathcal{G}\) can be instantiated by either a training cluster \(\mathcal{G}^{\mathrm{tr}}_k\) or an inference cluster \(\mathcal{G}^{\mathrm{inf}}_s\).

We further define the box associated with each cluster element as:
\begin{equation}
\rho_i
=
\begin{cases}
b_i, & i\in\mathcal{G}^{\mathrm{tr}}_k,\\
\mathrm{box}(d_i), & i\in\mathcal{G}^{\mathrm{inf}}_s,
\end{cases}
\end{equation}
where \(\rho_i\) denotes the spatial box used for region support construction. For training clusters, \(\rho_i\) is the dense proposal box \(b_i\). For inference clusters, \(\rho_i\) is the predicted box contained in proposal prediction \(d_i\).

For each constructed proposal cluster \(\mathcal{G}\), its spatial support region is defined as:
\begin{equation}
\mathcal{S}
=
\bigcup_{i\in\mathcal{G}} \rho_i,
\end{equation}
where \(\mathcal{S}\) corresponds to the spatial union of all proposal boxes within the cluster.

We further define the enclosing box:
\begin{equation}
\bar{b}
=
\mathrm{enclose}(\mathcal{S}),
\end{equation}
where \(\bar{b}\) is the minimal axis-aligned bounding box covering the entire support region. This design serves two purposes: (1) \(\mathcal{S}\) preserves the joint spatial support provided by multiple proposals, enabling a more complete representation of the object region; (2) \(\bar{b}\) provides a canonical spatial region for region-level feature extraction.

\subsubsection{Region-Level Semantic Evidence Extraction}

For each cluster, we crop the image patch using \(\bar{b}\) and apply masking using \(\mathcal{S}\). This operation preserves object-centric semantic regions while suppressing irrelevant background responses.

The resulting masked region observation is denoted as:
\begin{equation}
x_m
=
\mathrm{Mask}
\left(
\mathrm{Crop}(I,\bar{b}),\mathcal{S}
\right),
\end{equation}
where \(I\) denotes the input image, \(\mathrm{Crop}(I,\bar{b})\) extracts the patch within the enclosing box \(\bar{b}\), \(\mathrm{Mask}(\cdot,\mathcal{S})\) suppresses regions outside the cluster spatial support, and \(x_m\) denotes the canonical masked region observation constructed from cluster \(\mathcal{G}\).

We then extract the region-level semantic feature from the masked region
observation and map it into known-category logits:
\begin{equation}
\mathbf{f}
=
\phi(x_m),
\boldsymbol{\ell}
=
h_\theta(\mathbf{f}),
\end{equation}
where \(\phi(\cdot)\) denotes the feature encoder, e.g., a frozen DINOv3
encoder~\cite{dino_v3_2025}, \(\mathbf{f}\) denotes the semantic feature of
the constructed region, \(h_\theta(\cdot)\) denotes the semantic inference
head, and \(\boldsymbol{\ell}\in\mathbb{R}^{C}\) denotes the known-category
logits of cluster \(\mathcal{G}\).

\subsubsection{Semantic Consistency Inference}

To derive stable open-set semantic evidence, we first normalize logits:
\begin{equation}
\tilde{\boldsymbol{\ell}}
=
\alpha_l
\frac{
\boldsymbol{\ell}
}{
\|\boldsymbol{\ell}\|_2+\varepsilon
},
\end{equation}
where \(\tilde{\boldsymbol{\ell}}\) denotes the normalized logits, \(\alpha_l\) is a scaling factor, and \(\varepsilon\) ensures numerical stability. Logit normalization prevents unstable energy estimation caused by excessively large logit magnitudes, making semantic evidence more comparable across different regions.

To supervise region-consistent semantic inference, we optimize a dedicated region-level cross-entropy objective over the known categories on sampled training clusters:
\begin{equation}
\mathcal{L}_{\text{OCSC}}
=
-
\frac{1}{K_{\mathrm{tr}}}
\sum_{k=1}^{K_{\mathrm{tr}}}
\sum_{c=1}^{C}
\hat{y}_{k,c}
\log
\left(
\frac{
\exp(\tilde{\ell}^{\mathrm{tr}}_{k,c})
}{
\sum_{c'=1}^{C}
\exp(\tilde{\ell}^{\mathrm{tr}}_{k,c'})
}
\right),
\label{eq:ocsc_loss}
\end{equation}
where \(K_{\mathrm{tr}}\) denotes the number of sampled training clusters, \(\hat{y}_{k,c}\) is the one-hot known-category label of training cluster \(\mathcal{G}^{\mathrm{tr}}_k\), \(C\) denotes the number of known categories, and \(\tilde{\ell}^{\mathrm{tr}}_{k,c}\) denotes the normalized logit obtained by applying the shared region-evidence process to \(\mathcal{G}^{\mathrm{tr}}_k\) for category \(c\). The index \(c'\) in the denominator also enumerates known categories and is used only to avoid overloading the ground-truth index \(j\) in proposal assignment.

Based on the normalized logits, we compute the region-level knownness evidence score:
\begin{equation}
r
=
\log
\sum_{c=1}^{C}
\exp(\tilde{\ell}_{c}),
\end{equation}
where \(r\) quantifies the overall semantic support of the region representation within the known-category space, \(c\) indexes a known category, and \(C\) denotes the number of known categories.

Intuitively, a region corresponding to a known object category tends to activate at least one dominant known-category logit, resulting in a high knownness evidence score \(r\). In contrast, an unknown or semantically ambiguous region usually exhibits weak responses across all known categories, leading to a low evidence score. 

The latent semantic state of the region is therefore inferred as:
\begin{equation}
\hat{z}=
\begin{cases}
\arg\max_c \tilde{\ell}_{c}, & r\ge\delta, \\
\mathrm{unknown}, & r<\delta,
\end{cases}
\end{equation}
where \(\hat{z}\) denotes the inferred region-level semantic state, and \(\delta\) denotes the semantic decision threshold used to separate known-category evidence from unknown or semantically ambiguous regions.

Unlike conventional proposal-level open-set recognition, the proposed formulation performs semantic inference at the object-region level. Proposal predictions are thus conditioned on the inferred latent semantic state:
\begin{equation}
p(y_i \mid \rho_i,\hat{z}),
\qquad
i\in\mathcal{G},
\end{equation}
where \(y_i\) denotes the semantic prediction associated with the \(i\)-th proposal or proposal prediction.

Consequently, proposal predictions within the same object region are no longer treated as isolated decisions, but are jointly constrained by the inferred latent semantic state \(\hat{z}\). The inferred region semantics act as an object-centric prior that encourages correlated proposals to support a coherent known/unknown interpretation. When the region evidence strongly supports a known category, inconsistent unknown responses are suppressed; when the evidence shows insufficient compatibility with the known-category space, incompatible known-category predictions are suppressed, and the region is consistently treated as unknown. Thus, the inference process can be viewed as a structured approximation of the underlying object-centric semantic state rather than independent proposal-level classification.

As shown in Fig.~\ref{fig:inference}, the learned OCSC head is applied to each inference cluster \(\mathcal{G}^{\mathrm{inf}}_s\) to estimate its region-level semantic state \(\hat{z}_s\), which provides a shared semantic constraint for all proposal predictions in the cluster. Given the proposal prediction set \(\mathcal{D}=\{d_i\}_{i=1}^{n}\), where each \(d_i\) contains a predicted box \(\mathrm{box}(d_i)\) and a proposal-level semantic response, OCSC refines each prediction by enforcing its known/unknown response to be consistent with \(\hat{z}_s\):
\begin{equation}
\mathcal{D}'
=
\{d'_i\}_{i=1}^{n},
\qquad
d'_i
=
\mathrm{Refine}(d_i,\hat{z}_s),
\quad
i\in\mathcal{G}^{\mathrm{inf}}_s .
\end{equation}
Here, \(\mathcal{D}'\) denotes the region-consistent prediction set after OCSC refinement, and \(d'_i\) denotes the refined prediction corresponding to \(d_i\). The operator \(\mathrm{Refine}(d_i,\hat{z}_s)\) is a deterministic region-consistency filtering rule that suppresses proposal-level responses incompatible with the inferred cluster state: known-category predictions are suppressed when \(\hat{z}_s=\mathrm{unknown}\), while unknown responses are suppressed when \(\hat{z}_s\) indicates a known region. Predictions satisfying this semantic admissibility constraint retain their original boxes and confidence scores. Therefore, OCSC does not re-estimate object locations or introduce an additional detection branch; it reconciles proposal-level known/unknown responses under the shared region-level semantic state, converting \(\mathcal{D}\) into the region-consistent set \(\mathcal{D}'\).

\subsection{Geometry-Aware Boundary Uncertainty Mechanism}

Dense proposal mechanisms naturally generate many boundary-sensitive proposals around the same object. In OSOD, slight geometric perturbations of these proposals may directly trigger abrupt transitions between known and unknown predictions, leading to unstable open-set optimization.

To address this issue, we propose a \textbf{Geometry-Aware Boundary Uncertainty (GABU)} mechanism, which models geometry-induced semantic uncertainty and injects geometry-aware stabilization into the Unknown Probability Learner (UPL) of OpenDet~\citep{Han_2022_CVPR}. Instead of uniformly optimizing all proposals, GABU dynamically emphasizes geometrically ambiguous proposals near semantic transition boundaries, thereby facilitating more reliable semantic calibration and learning more stable open-set decision boundaries.

\subsubsection{Geometry-Induced Uncertainty Estimation and Reweighting}

To model this uncertainty, we define the localization quality of proposal \(b_i\) as:
\begin{equation}
u_i
=
\text{IoU}(b_i,g_j),
\end{equation}
where \(g_j\) denotes the matched target of proposal \(b_i\). Generally, low-IoU proposals tend to correspond to background regions, high-IoU proposals usually cover stable object centers, and intermediate-IoU proposals often lie near semantic transition regions and are therefore most vulnerable to known/unknown ambiguity.

\begin{figure}[!t]
    \centering
    \begin{overpic}[width=1\linewidth]{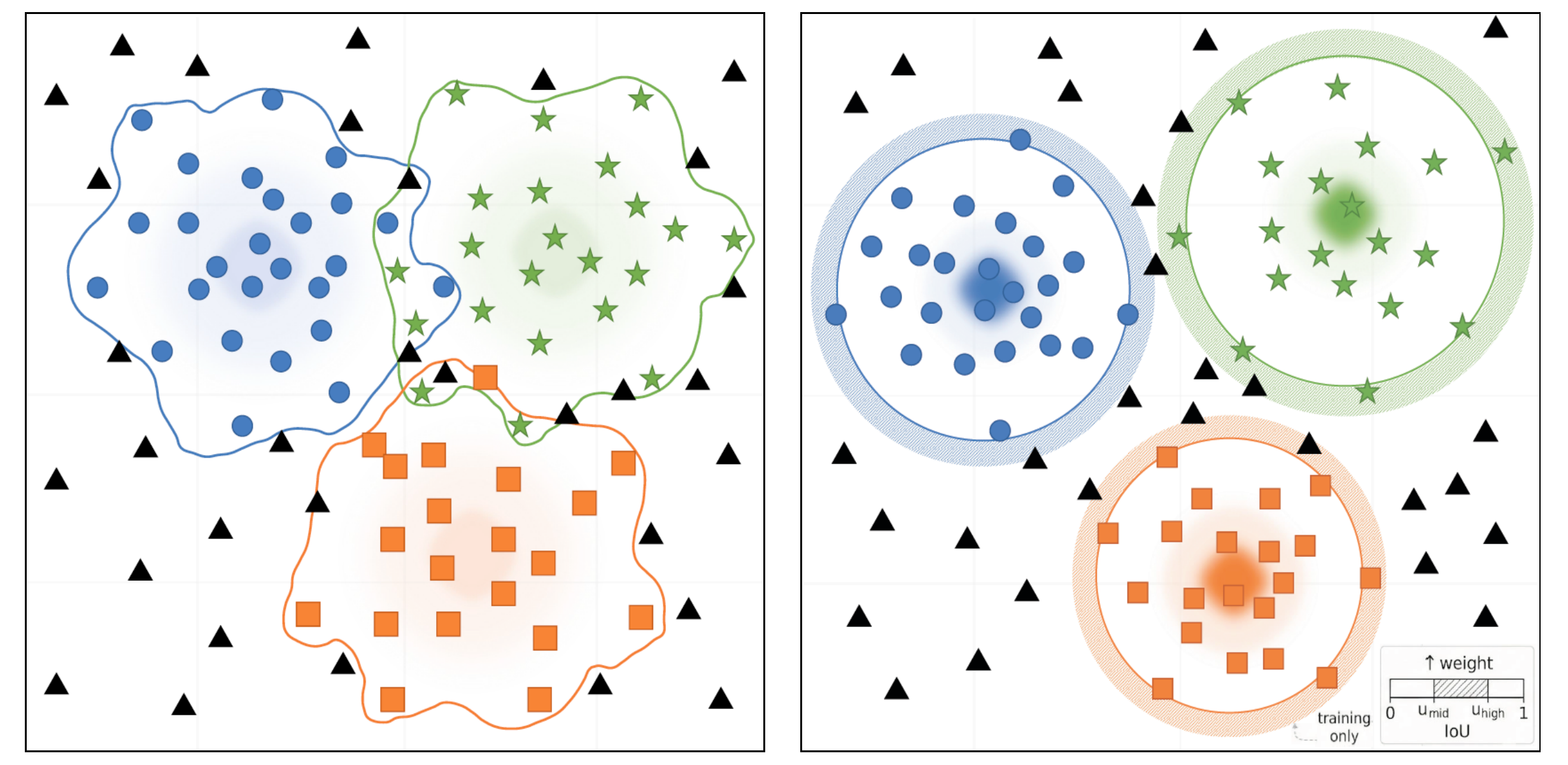}
    \put(12, -3){{\scriptsize (a) Constant weighting}}
    \put(60, -3){{\scriptsize (b) IoU-band reweighting}}
    \end{overpic}
    \vspace{0.05mm}
    \caption{Toy illustration of geometry-aware IoU-band reweighting for unknown-probability learning. Colored markers denote proposal features from different known classes, while black triangles indicate unknown proposals. (a) Constant per-proposal weighting treats overlaps equally. (b) Reweighting emphasizes the intermediate-IoU boundary band (inset), yielding more compact local supports and more stable open-set decisions.}
    \label{fig:gabu_schematic}
\end{figure}

As shown in \cref{fig:gabu_schematic}, we model geometry-induced uncertainty through an intermediate IoU band, where proposals near geometric decision boundaries receive additional emphasis in unknown-probability learning.

We build the uncertainty-aware optimization on the unknown-probability
learning (UPL) process~\citep{Han_2022_CVPR}, which provides the
proposal-level objective for estimating unknownness over mined RoIs. In the
base formulation, hard-example mining ranks candidate proposals by a mining
score \(m_i\). To align this selection with the geometry-aware boundary principle,
we attenuate the mining score for proposals that are both well localized and
highly confident as known:
\begin{equation}
\tilde{m}_i
=
m_i(1-u_i)^{\beta_m},
\quad
\text{if } u_i\ge u_{\text{high}} \text{ and } c_i\ge \tau,
\label{eq:gabu_mining}
\end{equation}
and set \(\tilde{m}_i=m_i\) otherwise. Here,
\(m_i\) denotes the original UPL mining score, \(u_i\) characterizes the localization quality of proposal \(b_i\),
\(u_{\text{high}}\) denotes the high-IoU boundary for well-localized proposals,
\(\tau\) denotes the known-confidence threshold, \(\mathcal{K}\) denotes the known-category index set, and
\(c_i=\max_{k\in\mathcal{K}}p_{i,k}\) denotes the maximum known-class
confidence computed from the predicted known-category probability \(p_{i,k}\) of proposal \(b_i\) for category \(k\). The coefficient \(\beta_m\) controls the attenuation strength. This adjustment
reduces the mining priority of easy high-IoU known proposals, preventing them
from dominating the mined set and shifting the selection focus toward
geometrically ambiguous boundary proposals.

Based on the adjusted mining scores, we obtain the mined foreground set
\(\mathcal{F}_{\text{mine}}\). For proposals in this set, geometry-induced
uncertainty is further incorporated by modulating the original UPL weighting term:
\begin{equation}
\omega_i
=
q_i(1-q_i)^{\alpha}
\left(
1+\gamma\,
\mathbb{I}
\left[
u_{\text{mid}}\le u_i < u_{\text{high}}
\right]
\right),
\quad i\in\mathcal{F}_{\text{mine}},
\label{eq:gabu_weight}
\end{equation}
where \(q_i\) represents the matched known-category confidence inherited from
the UPL weighting term, \(\alpha\) denotes the UPL weighting exponent, and \(\mathbb{I}[\cdot]\) is the indicator function. The coefficient \(\gamma>0\) controls the strength of
the geometry-aware scaling, and the interval
\([u_{\text{mid}},u_{\text{high}}]\) specifies the intermediate IoU band where
proposals are treated as geometrically ambiguous. The resulting geometry-aware
unknown-probability objective is:
\begin{equation}
\mathcal{L}_{\mathrm{UP}}^{\mathrm{GABU}}
=
\frac{1}{|\mathcal{F}_{\text{mine}}|}
\sum_{i\in\mathcal{F}_{\text{mine}}}
\omega_i
\ell_{\text{UP}}(s_i,y_i),
\label{eq:gabu_loss}
\end{equation}
where \(\ell_{\text{UP}}(\cdot)\) denotes the per-proposal
unknown-probability loss, \(s_i\) denotes the unknown-probability prediction, and \(y_i\) denotes the corresponding UPL training target. This modulation increases the contribution of
mined foreground proposals near the geometric boundary, where known/unknown
decisions are more likely to be unstable.


\subsection{Unified Optimization Objective}
\label{sec:objective}

The overall optimization objective of our method is defined as:
\begin{equation}
\mathcal{L}
=
\mathcal{L}^{\mathrm{GABU}}_{\text{base}}
+
\mathcal{L}_{\text{OCSC}},
\end{equation}
where \(\mathcal{L}^{\mathrm{GABU}}_{\text{base}}\) denotes the base detection objective inherited from our OpenDet-based implementation, with the original unknown-probability learning term replaced by the geometry-aware objective \(\mathcal{L}_{\mathrm{UP}}^{\mathrm{GABU}}\). It contains the standard proposal-level detection losses, the original open-set regularization terms except the original UPL term, the GABU-modified unknown-probability objective, and a lightweight prototype--text structuring regularizer:
\begin{equation}
\mathcal{L}^{\mathrm{GABU}}_{\text{base}}
=
\mathcal{L}_{\mathrm{OpenDet}}^{-\mathrm{UPL}}
+
\mathcal{L}_{\mathrm{UP}}^{\mathrm{GABU}}
+
\mathcal{L}_{\text{aux}}.
\end{equation}
Here, \(\mathcal{L}_{\mathrm{OpenDet}}^{-\mathrm{UPL}}\) denotes the OpenDet-based detection objective excluding the original UPL term, \(\mathcal{L}_{\text{aux}}\) summarizes auxiliary representation-structuring regularizers, including prototype pull regularization, prototype-cone unknown suppression, and text-anchored alignment, which are introduced as auxiliary regularization terms following existing prototype/center-based representation learning and CLIP-style text supervision~\citep{wen16centerloss,snell17protonets,radford21clip,cherti22scalingopenclip}, \(\mathcal{L}_{\text{OCSC}}\) supervises region-consistent semantic inference on sampled proposal clusters.

\section{Experiments} \label{sec:experiment}
\noindent


\begin{table*}[!t]
\centering
\caption{Main comparison results on four benchmarks. We report split-averaged $\mathrm{AP}_{\mathrm{known}}$, $\mathrm{AP}_{\mathrm{unk}}$, and the harmonic mean $\rm H$ (\%) for comparison. Results of competitors are from the published literature. The best results are marked in \textbf{bold}. }
\label{tab:osod3_main}
\label{tab:osod3_hmean}
\setlength{\tabcolsep}{2.2pt}
\renewcommand{\arraystretch}{1.05}
\resizebox{\textwidth}{!}{
\begin{tabular}{>{\kern-\tabcolsep}lccccccccccccc<{\kern-\tabcolsep}}
\toprule
& \multicolumn{3}{c}{OpenImages-Animal}
& \multicolumn{3}{c}{OpenImages-Vehicle}
& \multicolumn{3}{c}{CUB200}
& \multicolumn{3}{c}{MTSD}
& \multirow{2}{*}{Mean $\rm H$} \\
\cmidrule(lr){2-4}
\cmidrule(lr){5-7}
\cmidrule(lr){8-10}
\cmidrule(lr){11-13}
Method
& $\mathrm{AP}_{\mathrm{known}}$ & $\mathrm{AP}_{\mathrm{unk}}$ & $\rm H$
& $\mathrm{AP}_{\mathrm{known}}$ & $\mathrm{AP}_{\mathrm{unk}}$ & $\rm H$
& $\mathrm{AP}_{\mathrm{known}}$ & $\mathrm{AP}_{\mathrm{unk}}$ & $\rm H$
& $\mathrm{AP}_{\mathrm{known}}$ & $\mathrm{AP}_{\mathrm{unk}}$ & $\rm H$
& \\
\midrule
ORE~\citep{joseph21owod}
& 37.6 & 15.6 & 22.03
& 33.7 & 0.3 & 0.54
& 53.2 & 19.8 & 28.81
& 41.2 & 0.4 & 0.86
& 13.06 \\
DS~\citep{miller18dropout}
& 41.1 & 15.0 & 21.89
& 40.1 & 2.7 & 4.71
& 61.5 & 21.5 & 31.79
& 50.4 & 5.1 & 9.23
& 16.91 \\
VOS~\citep{du2022vos}
& 39.5 & 16.0 & 22.74
& 40.9 & 9.1 & 14.59
& 59.4 & 8.7 & 15.16
& 49.1 & 4.7 & 8.46
& 15.24 \\
Orth.Det~\citep{Sun_2024_CVPR}
& \textbf{44.8} & 8.9 & 14.13
& 40.5 & 5.6 & 8.17
& \textbf{66.5} & 24.7 & 35.92
& 49.3 & 3.3 & 6.07
& 16.07 \\
FCOS~\citep{tian2019fcos}
& 30.3 & \textbf{41.8} & 34.82
& 30.7 & 18.7 & 21.31
& 53.5 & 24.7 & 33.74
& 41.7 & 4.4 & 7.96
& 24.46 \\
Faster R-CNN~\citep{ren2015faster}
& 37.8 & 35.3 & 36.42
& 39.9 & 17.0 & 22.84
& 62.2 & 24.2 & 34.75
& 50.0 & 3.1 & 5.75
& 24.94 \\
\midrule
OpenDet~\citep{Han_2022_CVPR}
& 36.9 & 33.0 & 34.76
& 38.7 & 14.4 & 20.29
& 63.3 & 27.0 & 37.76
& \textbf{51.8} & 9.9 & 16.35
& 27.29 \\
\rowcolor{blue!10}
OCSI-Net (Ours)
& 42.2 & 38.6 & \textbf{40.27}
& \textbf{42.5} & \textbf{19.8} & \textbf{26.42}
& 63.4 & \textbf{40.5} & \textbf{49.30}
& 51.3 & \textbf{13.0} & \textbf{20.47}
& \textbf{34.12} \\
\bottomrule
\end{tabular}
}
\end{table*}

\begin{table*}[!t]
\caption{Split-wise OSOD-III \cite{hosoya25rethinkingosod} results (\%) on OpenImages superclasses. We report $\mathrm{AP}_{\mathrm{known}}$, $\mathrm{AP}_{\mathrm{unk}}$, and their $\rm H$.}
\label{tab:app_openimages_splits}
\centering
\setlength{\tabcolsep}{1.35pt}
\renewcommand{\arraystretch}{1.05}
\resizebox{\textwidth}{!}{%
\begin{tabular}{>{\kern-\tabcolsep}l*{15}{c}<{\kern-\tabcolsep}}
\toprule
& \multicolumn{3}{c}{Split1} & \multicolumn{3}{c}{Split2} & \multicolumn{3}{c}{Split3} & \multicolumn{3}{c}{Split4} & \multicolumn{3}{c}{Mean} \\
\cmidrule(lr){2-4}\cmidrule(lr){5-7}\cmidrule(lr){8-10}\cmidrule(lr){11-13}\cmidrule(lr){14-16}
Method
& $\mathrm{AP}_{\mathrm{known}}$ & $\mathrm{AP}_{\mathrm{unk}}$ & $\rm H$
& $\mathrm{AP}_{\mathrm{known}}$ & $\mathrm{AP}_{\mathrm{unk}}$ & $\rm H$
& $\mathrm{AP}_{\mathrm{known}}$ & $\mathrm{AP}_{\mathrm{unk}}$ & $\rm H$
& $\mathrm{AP}_{\mathrm{known}}$ & $\mathrm{AP}_{\mathrm{unk}}$ & $\rm H$
& $\mathrm{AP}_{\mathrm{known}}$ & $\mathrm{AP}_{\mathrm{unk}}$ & $\rm H$ \\
\midrule
\multicolumn{16}{c}{\textbf{OpenImages-Animal}} \\
\midrule
ORE~\citep{joseph21owod} & 40.4 & 17.4 & 24.32 & 34.8 & 13.0 & 18.93 & 40.4 & 19.1 & 25.94 & 34.8 & 13.0 & 18.93 & 37.6 & 15.6 & 22.03 \\
DS~\citep{miller18dropout} & 44.0 & 19.0 & 26.54 & 36.8 & 12.3 & 18.44 & 43.3 & 14.0 & 21.16 & 40.2 & 14.6 & 21.42 & 41.1 & 15.0 & 21.89 \\
VOS~\citep{du2022vos} & 39.5 & 17.5 & 24.25 & 37.5 & 13.9 & 20.28 & 43.1 & 14.7 & 21.92 & 37.9 & 18.1 & 24.50 & 39.5 & 16.0 & 22.74 \\
Orth.Det~\citep{Sun_2024_CVPR} & \textbf{47.0} & 11.8 & 18.86 & \textbf{40.5} & 4.7 & 8.42 & \textbf{47.2} & 3.1 & 5.82 & \textbf{44.5} & 15.9 & 23.43 & \textbf{44.8} & 8.9 & 14.13 \\
FCOS~\citep{tian2019fcos} & 35.0 & \textbf{44.4} & 39.14 & 30.8 & \textbf{35.6} & 33.03 & 32.6 & \textbf{43.7} & 37.34 & 22.6 & \textbf{43.6} & 29.77 & 30.3 & \textbf{41.8} & 34.82 \\
Faster R-CNN~\citep{ren2015faster} & 41.8 & 36.9 & 39.20 & 34.0 & 29.5 & 31.59 & 39.7 & 37.7 & 38.67 & 35.5 & 37.0 & 36.23 & 37.8 & 35.3 & 36.42 \\
\midrule
OpenDet~\citep{Han_2022_CVPR} & 42.4 & 34.9 & 38.29 & 23.2 & 25.8 & 24.43 & 43.0 & 37.9 & 40.29 & 39.0 & 33.5 & 36.04 & 36.9 & 33.0 & 34.76 \\
\rowcolor{blue!10} OCSI-Net (Ours) & 45.1 & 39.6 & \textbf{42.17} & 36.5 & 33.0 & \textbf{34.66} & 46.3 & 42.3 & \textbf{44.21} & 40.8 & 39.3 & \textbf{40.04} & 42.2 & 38.6 & \textbf{40.27} \\
\midrule
\multicolumn{16}{c}{\textbf{OpenImages-Vehicle}} \\
\midrule
ORE~\citep{joseph21owod} & 46.9 & 0.5 & 0.99 & 35.0 & 0.1 & 0.20 & 25.0 & 0.2 & 0.40 & 27.7 & 0.3 & 0.59 & 33.7 & 0.3 & 0.54 \\
DS~\citep{miller18dropout} & 52.6 & 0.5 & 0.99 & 40.7 & 2.3 & 4.35 & 31.9 & 6.5 & 10.80 & 35.1 & 1.4 & 2.69 & 40.1 & 2.7 & 4.71 \\
VOS~\citep{du2022vos} & \textbf{53.2} & 7.4 & 12.99 & 41.9 & 7.1 & 12.14 & 32.8 & 9.4 & 14.61 & 35.7 & 12.6 & 18.63 & 40.9 & 9.1 & 14.59 \\
Orth.Det~\citep{Sun_2024_CVPR} & \textbf{56.3} & 2.5 & 4.79 & 41.3 & 1.2 & 2.33 & 29.7 & 16.1 & 20.88 & 34.6 & 2.5 & 4.66 & 40.5 & 5.6 & 8.17 \\
FCOS~\citep{tian2019fcos} & 49.6 & 14.2 & 22.08 & 32.7 & 14.6 & 20.19 & 19.2 & \textbf{24.7} & 21.61 & 21.4 & 21.3 & 21.35 & 30.7 & 18.7 & 21.31 \\
Faster R-CNN~\citep{ren2015faster} & 51.0 & 10.5 & 17.41 & 42.0 & \textbf{15.2} & 22.32 & 31.0 & 22.1 & 25.80 & 35.7 & 20.2 & 25.80 & 39.9 & 17.0 & 22.84 \\
\midrule
OpenDet~\citep{Han_2022_CVPR} & 50.6 & 10.2 & 16.98 & 40.4 & 12.5 & 19.09 & 30.2 & 15.9 & 20.83 & 33.6 & 19.0 & 24.27 & 38.7 & 14.4 & 20.29 \\
\rowcolor{blue!10} OCSI-Net (Ours) & 53.1 & \textbf{18.4} & \textbf{27.33} & \textbf{43.6} & 14.7 & \textbf{21.99} & \textbf{34.4} & 22.9 & \textbf{27.50} & \textbf{38.8} & \textbf{23.0} & \textbf{28.88} & \textbf{42.5} & \textbf{19.8} & \textbf{26.42} \\
\bottomrule
\end{tabular}}
\end{table*}

\begin{table*}[!t]
\caption{Split-wise OSOD-III \cite{hosoya25rethinkingosod} results (\%) on CUB200. We report $\mathrm{AP}_{\mathrm{known}}$, $\mathrm{AP}_{\mathrm{unk}}$, and their $\rm H$ on four official CUB200 splits.}
\label{tab:app_cub_splits}
\centering
\setlength{\tabcolsep}{1.45pt}
\renewcommand{\arraystretch}{1.05}
\resizebox{\textwidth}{!}{%
\begin{tabular}{>{\kern-\tabcolsep}l*{15}{c}<{\kern-\tabcolsep}}
\toprule
& \multicolumn{3}{c}{Split1} & \multicolumn{3}{c}{Split2} & \multicolumn{3}{c}{Split3} & \multicolumn{3}{c}{Split4} & \multicolumn{3}{c}{Mean} \\
\cmidrule(lr){2-4}\cmidrule(lr){5-7}\cmidrule(lr){8-10}\cmidrule(lr){11-13}\cmidrule(lr){14-16}
Method
& $\mathrm{AP}_{\mathrm{known}}$ & $\mathrm{AP}_{\mathrm{unk}}$ & $\rm H$
& $\mathrm{AP}_{\mathrm{known}}$ & $\mathrm{AP}_{\mathrm{unk}}$ & $\rm H$
& $\mathrm{AP}_{\mathrm{known}}$ & $\mathrm{AP}_{\mathrm{unk}}$ & $\rm H$
& $\mathrm{AP}_{\mathrm{known}}$ & $\mathrm{AP}_{\mathrm{unk}}$ & $\rm H$
& $\mathrm{AP}_{\mathrm{known}}$ & $\mathrm{AP}_{\mathrm{unk}}$ & $\rm H$ \\
\midrule
ORE~\citep{joseph21owod} & 51.3 & 18.1 & 26.76 & 53.6 & 21.8 & 30.99 & 54.4 & 17.7 & 26.71 & 53.6 & 21.6 & 30.79 & 53.2 & 19.8 & 28.81 \\
DS~\citep{miller18dropout} & 61.7 & 19.6 & 29.75 & 61.2 & 22.2 & 32.58 & 62.8 & 22.2 & 32.80 & 60.4 & 21.8 & 32.04 & 61.5 & 21.5 & 31.79 \\
VOS~\citep{du2022vos} & 59.7 & 8.1 & 14.26 & 59.5 & 9.1 & 15.79 & 60.5 & 8.1 & 14.29 & 57.7 & 9.5 & 16.31 & 59.4 & 8.7 & 15.16 \\
Orth.Det~\citep{Sun_2024_CVPR} & \textbf{64.9} & 27.2 & 38.33 & \textbf{67.2} & 26.9 & 38.42 & \textbf{67.7} & 20.9 & 31.94 & \textbf{66.0} & 23.8 & 34.98 & \textbf{66.5} & 24.7 & 35.92 \\
FCOS~\citep{tian2019fcos} & 55.0 & 23.0 & 32.44 & 55.2 & 26.1 & 35.44 & 50.6 & 25.0 & 33.47 & 53.0 & 24.6 & 33.60 & 53.5 & 24.7 & 33.74 \\
Faster R-CNN~\citep{ren2015faster} & 62.0 & 21.6 & 32.04 & 62.7 & 26.2 & 36.96 & 63.2 & 24.0 & 34.79 & 60.8 & 24.8 & 35.23 & 62.2 & 24.2 & 34.75 \\
\midrule
OpenDet~\citep{Han_2022_CVPR} & 63.9 & 23.1 & 33.93 & 63.6 & 30.0 & 40.77 & 63.9 & 26.3 & 37.26 & 61.6 & 28.6 & 39.06 & 63.3 & 27.0 & 37.76 \\
\rowcolor{blue!10} OCSI-Net (Ours) & 63.8 & \textbf{35.1} & \textbf{45.29} & 62.9 & \textbf{45.2} & \textbf{52.60} & 64.3 & \textbf{41.2} & \textbf{50.22} & 62.6 & \textbf{40.4} & \textbf{49.11} & 63.4 & \textbf{40.5} & \textbf{49.30} \\
\bottomrule
\end{tabular}}
\end{table*}

\begin{table*}[!t]
\caption{Split-wise OSOD-III \cite{hosoya25rethinkingosod} results (\%) on MTSD. Following the benchmark protocol, we report $\mathrm{AP}_{\mathrm{known}}$ on the known set (K) and $\mathrm{AP}_{\mathrm{unk}}$ on three unknown evaluation sets (U1, U2, and U1+2), and their $\rm H$. }
\label{tab:app_mtsd_splits}
\centering
\scriptsize
\setlength{\tabcolsep}{4.0pt} 
\renewcommand{\arraystretch}{1}
\resizebox{\textwidth}{!}{
\begin{tabular}{>{\kern-\tabcolsep}lccccccccc<{\kern-\tabcolsep}}
\toprule
Method
& K ($\mathrm{AP}_{\mathrm{known}}$)
& U1 ($\mathrm{AP}_{\mathrm{unk}}$) & $\rm H$
& U2 ($\mathrm{AP}_{\mathrm{unk}}$) & $\rm H$
& U1+2 ($\mathrm{AP}_{\mathrm{unk}}$) &$\rm H$
& Mean ($\mathrm{AP}_{\mathrm{unk}}$)
& Mean ($\rm H$) \\
\midrule
ORE~\citep{joseph21owod} & 41.2 & 0.4 & 0.79 & 0.2 & 0.40 & 0.7 & 1.38 & 0.4 & 0.86 \\
DS~\citep{miller18dropout} & 50.4 & 4.5 & 8.26 & 3.4 & 6.37 & 7.5 & 13.06 & 5.1 & 9.23 \\
VOS~\citep{du2022vos} & 49.1 & 4.6 & 8.41 & 2.9 & 5.48 & 6.5 & 11.48 & 4.7 & 8.46 \\
Orth.Det~\citep{Sun_2024_CVPR} & 49.3 & 3.9 & 7.23 & 1.5 & 2.91 & 4.4 & 8.08 & 3.3 & 6.07 \\
FCOS~\citep{tian2019fcos} & 41.7 & 3.8 & 6.97 & 3.3 & 6.12 & 6.2 & 10.79 & 4.4 & 7.96 \\
Faster R-CNN~\citep{ren2015faster} & 50.0 & 2.5 & 4.76 & 2.3 & 4.40 & 4.4 & 8.09 & 3.1 & 5.75 \\
\midrule
OpenDet~\citep{Han_2022_CVPR} & \textbf{51.8} & 8.7 & 14.90 & 6.7 & 11.87 & 14.2 & 22.29 & 9.9 & 16.35 \\
\rowcolor{blue!10} OCSI-Net (Ours) & 51.3 & \textbf{12.9} & \textbf{20.62} & \textbf{8.2} & \textbf{14.14} & \textbf{18.0} & \textbf{26.65} & \textbf{13.0} & \textbf{20.47} \\
\bottomrule
\end{tabular}}
\end{table*}

\subsection{Experimental Setup}

We evaluate our method on three benchmarks ranging from large-scale natural images, fine-grained to object-specific scenarios: OpenImages \citep{kuznetsova2020openimages}, CUB-200-2011 \citep{wah2011caltech}, and MTSD \citep{neuhold2017mtsd}. In the adopted OSOD-III \cite{hosoya25rethinkingosod} evaluation protocol, OpenImages is evaluated on two superclasses, i.e., Animal and Vehicle. Training is conducted using only the designated known categories, while evaluation additionally includes unknown categories that are disjoint from the known label space. All experiments strictly follow the official benchmark protocols \citep{hosoya25rethinkingosod}, including the predefined known/unknown splits and evaluation criteria.

 We use SGD with momentum $0.9$ and weight decay $1\times10^{-4}$, together with a two-step learning-rate decay, to optimize the model. For both OpenImages and CUB200 datasets, we use multi-scale training by uniformly sampling the shorter side from 480 to 800 in increments of 32, and perform single-scale inference with the shorter side fixed to 800. For the MTSD dataset, we adopt an enlarged scale range from 960 to 1600 (step 64) and test at a single scale of 1600, with the maximum size set to 2666.


OCSC trains the head on frozen DINOv3 features \citep{dino_v3_2025}, where region crops are resized to $224\times224$. Training clusters are constructed using
\(\tau_{\mathrm{clus}}=0.6\), which is also adopted as the inference grouping threshold \(\tau_{\mathrm{grp}}\). At inference time, OCSC uses a fixed split-specific semantic decision threshold $\delta$. For GABU, we set \(u_{\text{mid}}=0.3\), \(u_{\text{high}}=0.6\), and the uncertainty scaling factor \(\gamma=1.2\).

\subsection{Comparison Results}
\label{subsec:benchmark_results}

\subsubsection{Overall Comparison}
\label{sec:overall_comparison}

We compare our method against the recent OSOD competitors, including ORE~\citep{joseph21owod},
Dropout Sampling (DS)~\citep{miller18dropout},
VOS~\citep{du2022vos},
Orth.Det~\citep{Sun_2024_CVPR},
FCOS~\citep{tian2019fcos},
Faster R-CNN~\citep{ren2015faster},
and the baseline OpenDet~\citep{Han_2022_CVPR}. 


As shown in Table \ref{tab:osod3_main}, OCSI-Net demonstrates strong and consistent performance across all four benchmarks. In terms of $\mathrm{AP}_{\mathrm{known}}$, OCSI-Net attains competitive results on all datasets, achieving the best performance on OpenImages-Vehicle and comparable to the best methods on both CUB200 and MTSD datasets. Besides, OCSI-Net consistently achieves the highest $\mathrm{AP}_{\mathrm{unk}}$ 
on three out of four benchmarks, with particularly significant margins on both OpenImages-Animal and CUB200. These results underscore its superior capability in discovering and accurately localizing previously unseen objects.


To better characterize the trade-off between known-class detection and unknown-object discovery, we further report a harmonic-mean metric between $\mathrm{AP}_{\mathrm{known}}$ and $\mathrm{AP}_{\mathrm{unk}}$. Specifically, the harmonic-mean metric is formulated as:
\begin{equation}
 \rm H =\; \frac{2 \times \mathrm{AP}_{\mathrm{known}}\times \mathrm{AP}_{\mathrm{unk}}}{\mathrm{AP}_{\mathrm{known}} + \mathrm{AP}_{\mathrm{unk}}}.
\end{equation}

Table~\ref{tab:osod3_hmean} highlights the comparison results of our method and the competitors in terms of $\rm H$. From the results, OCSI-Net consistently achieves the best H across all four datasets, outperforming the second-best method by an average of 6.83\%, which demonstrates a superior balance between known-class detection and unknown-object discovery. 

\begin{table*}[!t]
\centering
\caption{Ablation results (\%) on the CUB200 dataset. We report split-wise $\mathrm{AP}_{\mathrm{known}}$ and $\mathrm{AP}_{\mathrm{unk}}$, as well as their harmonic mean $\rm H$. 
}
\label{tab:ablation}
\resizebox{0.97\linewidth}{!}{
\begin{tabular}{cc|ccc|ccc|ccc|ccc}
\toprule
  \multicolumn{2}{c|}{\textbf{Settings}} &
  \multicolumn{3}{c|}{\textbf{Split 1}} &
  \multicolumn{3}{c|}{\textbf{Split 2}} &
  \multicolumn{3}{c|}{\textbf{Split 3}} &
  \multicolumn{3}{c}{\textbf{Split 4}} \\
  \midrule
  GABU & OCSC & $\mathrm{AP}_{\mathrm{known}}$ &
  $\mathrm{AP}_{\mathrm{unk}}$ &
  $\rm H$ &
  $\mathrm{AP}_{\mathrm{known}}$  &
  $\mathrm{AP}_{\mathrm{unk}}$ &
  $\rm H$ &
  $\mathrm{AP}_{\mathrm{known}}$  &
  $\mathrm{AP}_{\mathrm{unk}}$ &
  $\rm H$ &
  $\mathrm{AP}_{\mathrm{known}}$  &
  $\mathrm{AP}_{\mathrm{unk}}$ &
  $\rm H$  \\ 
  \midrule
   \ding{55}  & \ding{55} & \textbf{63.9} & 23.1 & 33.93 & 63.6 & 30.0 & 40.80 & 63.9 & 26.3 & 37.26 & 61.6 & 28.6 & 39.06 \\
\checkmark & \ding{55}  & 63.3 & 30.0 & 40.71
& 62.5 & 39.1 & 48.11
& 64.1 & 34.6 & 44.94
& 62.1 & 37.5 & 46.76 \\ 
\ding{55} & \checkmark & 63.0 & 29.5 & 40.18
& \textbf{63.8} & 37.4 & 47.16
& \textbf{64.7} & 34.0 & 44.58
& \textbf{63.0} & 34.1 & 44.25
\\ 
\checkmark & \checkmark & 63.8 & \textbf{35.1} & \textbf{45.29}
& 62.9 & \textbf{45.2} & \textbf{52.60}
& 64.3 & \textbf{41.2} & \textbf{50.22}
& 62.6 & \textbf{40.4} & \textbf{49.11} \\ 
\bottomrule
\end{tabular}}
\end{table*}

\subsubsection{Split-wise Results}
\label{sec:splitwise_results}

To evaluate robustness under different label partitions, we perform a split-wise analysis on all datasets with official splits. Since the composition of unknown classes and their semantic similarity to known categories may vary substantially across splits, per-split evaluation is essential for a fair assessment of open-set generalization. We therefore report $\mathrm{AP}_{\mathrm{known}}$ and $\mathrm{AP}_{\mathrm{unk}}$, and $\rm H$ for each split.

For the MTSD dataset, we follow the benchmark evaluation protocol by reporting $\mathrm{AP}_{\mathrm{known}}$ on the known set (K) and $\mathrm{AP}_{\mathrm{unk}}$ on three unknown evaluation sets (U1, U2, and U1+2). In this case, the summary mean is computed over the three unknown evaluation sets \{U1, U2, U1+2\}, and we likewise report the harmonic mean between K and each unknown set to quantify the trade-off. All comparison results are directly from \citet{hosoya25rethinkingosod}.

From Table~\ref{tab:app_openimages_splits}, Table~\ref{tab:app_cub_splits}, and Table~\ref{tab:app_mtsd_splits}, the split-wise results indicate that the proposed method improves unknown detection consistently across splits, while maintaining competitive known-set accuracy. Notably, the improvements remain consistent across different known/unknown partitions, suggesting that the gains are not driven by a single favorable split but persist under different known/unknown partitions. Therefore, these split-wise results further support the conclusions in Table~\ref{tab:osod3_main} and validate the stability of our method under diverse known/unknown partition settings.


\subsection{Ablation Study}
\label{subsec:ablation_cub}

In this section, we conduct ablation experiments on CUB200 to evaluate the impacts of OCSC and GABU.

\subsubsection{Main Component Ablation}
\label{sec:ablation_components}

Table~\ref{tab:ablation} presents the ablation results on CUB200, highlighting both the individual contributions of each component and their complementary effects. Starting from the baseline, incorporating either GABU or OCSC consistently improves $\mathrm{AP}_{\mathrm{unk}}$ and the harmonic mean $\mathrm{H}$ across all splits, indicating that both components independently enhance unknown-category detection while largely preserving known-category performance. GABU generally yields larger gains in $\mathrm{AP}_{\mathrm{unk}}$, suggesting its effectiveness in promoting unknown-object discovery, whereas OCSC leads to more balanced improvements and occasionally achieves slightly higher $\mathrm{AP}_{\mathrm{known}}$, reflecting its role in stabilizing known-class detection. When both components are combined in OCSI-Net, the model achieves the best performance across all splits, with substantial improvements in $\mathrm{AP}_{\mathrm{unk}}$ and the highest harmonic mean in every case. Notably, OCSI-Net consistently improves $\mathrm{H}$ by a large margin over the baseline while maintaining comparable $\mathrm{AP}_{\mathrm{known}}$, demonstrating that the full model effectively balances known and unknown detection. Overall, these results confirm that GABU and OCSC are complementary and that their joint integration is crucial for achieving robust and well-balanced open-set detection performance.

\subsubsection{GABU Hyperparameter Sensitivity}
\label{sec:ablation_gabu_hyper}

We further study the sensitivity of GABU to its key reweighting hyperparameters on CUB200 Split-2. Unless otherwise specified, we use the default overlap-driven reweighting configuration with midpoint $u_{\text{mid}}=0.3$, highpoint $u_{\text{high}}=0.6$, and scaling factor $\gamma=1.2$. Table~\ref{tab:gabu_ablation} reports $\mathrm{AP}_{\mathrm{known}}$, $\mathrm{AP}_{\mathrm{unk}}$, and their $\rm H$.

\begin{table}[!t]
\centering
\caption{GABU hyperparameter ablations (\%) on CUB200 Split-2. We vary one parameter at a time around the default setting (Base). $\rm H$ denotes the harmonic mean of $\mathrm{AP}_{\mathrm{known}}$ and $\mathrm{AP}_{\mathrm{unk}}$.}
\label{tab:gabu_ablation}
\setlength{\tabcolsep}{6pt}
\renewcommand{\arraystretch}{1.15}
\sisetup{
  detect-weight=true,
  detect-family=true
}

\begin{tabular}{
  >{\kern-\tabcolsep}c !{\vrule width 0.6pt}
  >{\small}c
  >{\small}S[table-format=2.1]
  >{\small}S[table-format=2.1]
  >{\small}S[table-format=2.2]<{\kern-\tabcolsep}
}
\toprule
\textbf{Param.} & \multicolumn{1}{c}{\textbf{Value}} &
\multicolumn{1}{c}{$\mathrm{AP}_{\mathrm{known}}$} &
\multicolumn{1}{c}{$\mathrm{AP}_{\mathrm{unk}}$} &
\multicolumn{1}{c}{$\rm H$} \\
\midrule
\multirow{2}{*}{\large$\displaystyle u_{\text{high}}$} & 0.5 & \bfseries 63.8 & 44.7 & 52.57 \\
                                  & 0.7 & 62.6 & 42.8 & 50.84 \\
\midrule
\multirow{2}{*}{\large$\displaystyle u_{\text{mid}}$}  & 0.2 & 63.1 & 43.8 & 51.71 \\
                                  & 0.4 & 63.6 & 43.8 & 51.87 \\
\midrule
\multirow{2}{*}{\large$\displaystyle \gamma$}          & 1.0 & 63.3 & 44.4 & 52.19 \\
                                  & 1.4 & 63.2 & 42.2 & 50.61 \\
\midrule
\rowcolor{blue!10}
\textbf{Base (default)} & {} & 62.9 & \bfseries 45.2 & \bfseries 52.60 \\
\bottomrule
\end{tabular}
\end{table}

\begin{figure*}[!t]
\centering
\begin{subfigure}[!t]{0.49\textwidth}
\centering
\includegraphics[width=\linewidth]{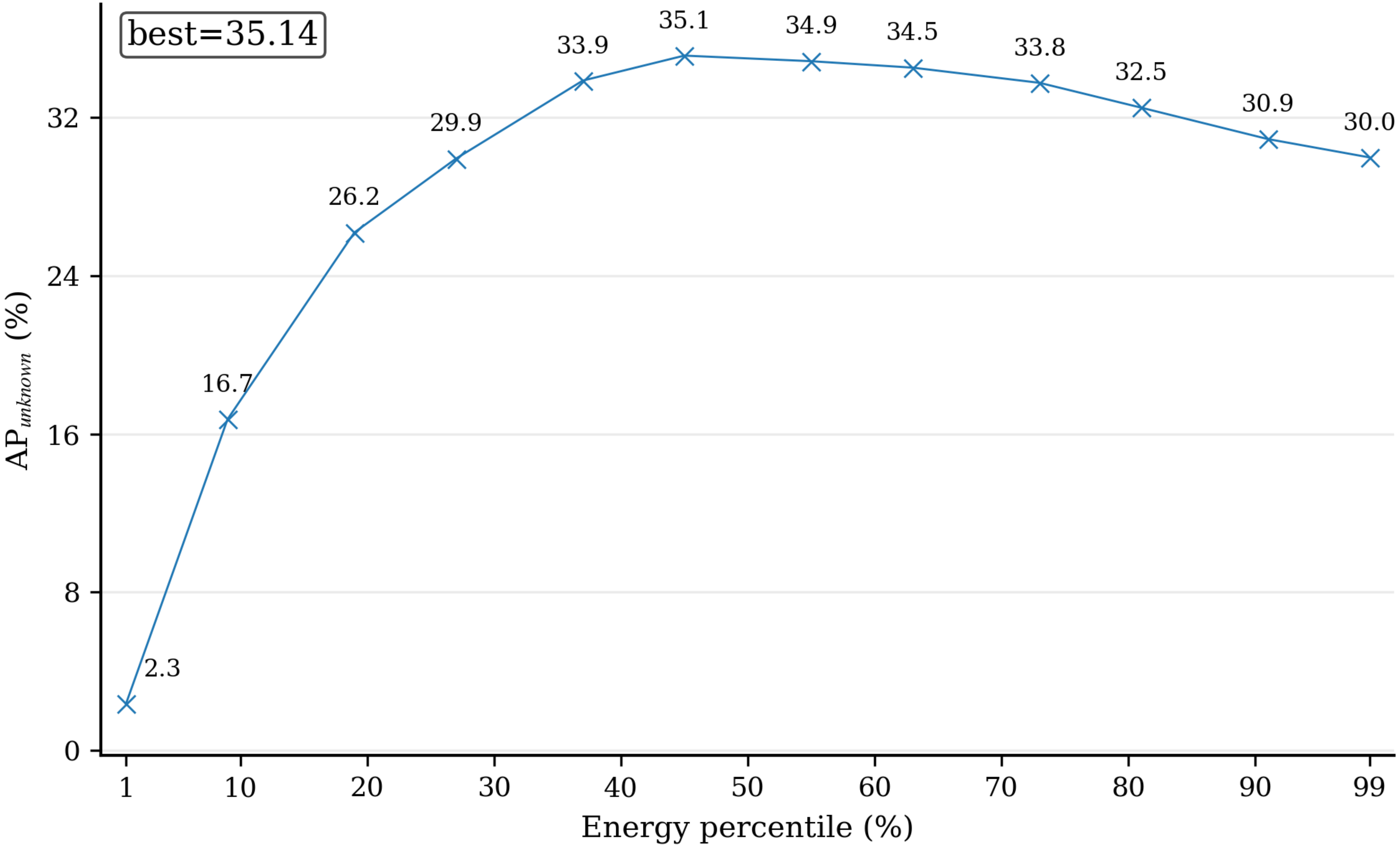}
\caption{Split1}
\end{subfigure}
\hfill
\begin{subfigure}[!t]{0.49\textwidth}
\centering
\includegraphics[width=\linewidth]{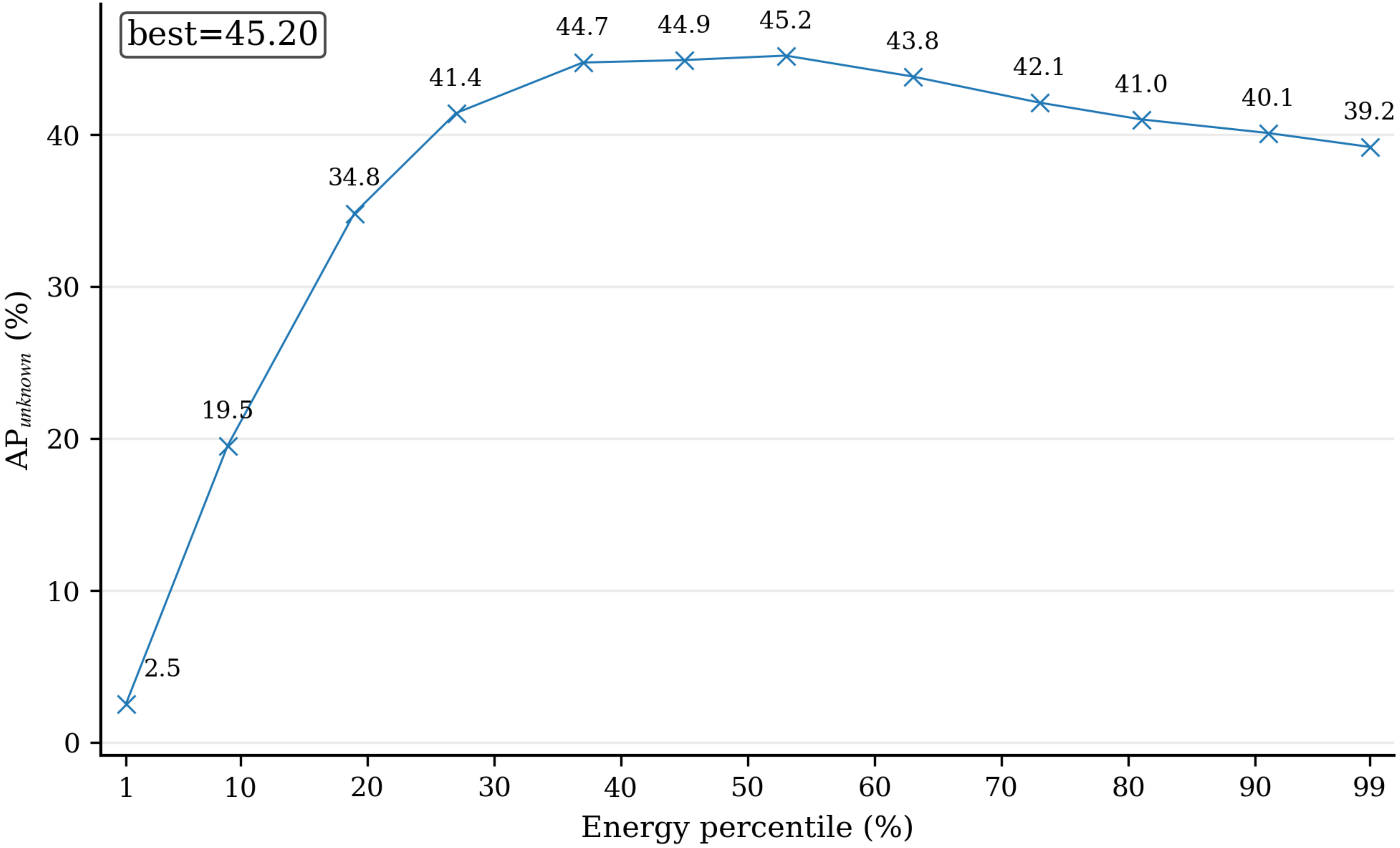}
\caption{Split2}
\end{subfigure}
\caption{Sensitivity of the OCSC semantic decision threshold $\delta$ on CUB200. On two CUB200 splits, we vary $\delta$ by setting it to the $p$-th percentile of the region-level knownness evidence scores $\{r_s\}$ computed on the evaluated split.}
\label{fig:cub_delta_ablation}
\vspace{-1.5mm}
\end{figure*}

Overall, the default configuration achieves the best $\mathrm{AP}_{\mathrm{unk}}$ and the best harmonic mean, indicating that the chosen overlap regime and reweighting strength provide a favorable balance between known-category accuracy and unknown-object discovery. Reducing $u_{\text{high}}$ to $0.5$ slightly increases $\mathrm{AP}_{\mathrm{known}}$ but decreases $\mathrm{AP}_{\mathrm{unk}}$, suggesting that an overly permissive highpoint weakens the emphasis on geometrically ambiguous samples that are most beneficial for unknown detection. In contrast, increasing $u_{\text{high}}$ to $0.7$ degrades both metrics, consistent with an overly conservative highpoint that under-emphasizes boundary-band proposals and yields less effective stabilization.

Varying $u_{\text{mid}}$ within $\{0.2,0.4\}$ leads to modest changes in $\mathrm{AP}_{\mathrm{known}}$ but consistently lower $\mathrm{AP}_{\mathrm{unk}}$ than the default, highlighting that the midpoint primarily controls where the reweighting concentrates along the IoU spectrum. Finally, the scaling factor $\gamma$ exhibits a clear trade-off: $\gamma=1.0$ remains competitive, while a larger $\gamma=1.4$ notably hurts $\mathrm{AP}_{\mathrm{unk}}$, implying that overly aggressive amplification can over-correct training and reduce open-set reliability.

\subsubsection{OCSC Semantic Decision Threshold Sensitivity}
\label{sec:ablation_ocsc_delta}

In this subsection, we take CUB200 as a representative example to study the effect of the OCSC semantic decision rule, with a particular focus on the sensitivity to the threshold $\delta$. To more clearly present the effect of the threshold $\delta$, we conduct a sensitivity analysis over $\delta$ and report the corresponding performance curves under different threshold values (Fig.~\ref{fig:cub_delta_ablation}). Specifically, for each evaluated split, we compute the set of region-level knownness evidence scores $\{r_s\}$ produced on the test set and set $\delta$ to the $p$-th percentile of these scores, where $p$ is swept over a range of percentiles. Intuitively, a smaller $p$ yields a more permissive semantic decision, whereas a larger $p$ yields a more conservative semantic decision. This sweep is used only to visualize how the refinement behavior changes with $\delta$; in the main results, $\delta$ is fixed within each split.

Fig.~\ref{fig:cub_delta_ablation} reports the resulting $\mathrm{AP}_{\mathrm{unk}}$ as a function of the percentile $p$ on the two official CUB200 splits. Across splits, overly permissive thresholds tend to under-reject ambiguous regions and thus hurt unknown detection, while overly conservative thresholds can also degrade $\mathrm{AP}_{\mathrm{unk}}$ by inducing excessive unknown responses and reducing the reliability of region-level refinement. The best performance is typically attained at an intermediate percentile, indicating that OCSC benefits from a calibrated balance between accepting confident known regions and rejecting ambiguous regions as unknown. Overall, this ablation suggests that OCSC is not brittle to the exact choice of $\delta$: the curves exhibit a broad high-performing regime around intermediate percentiles, supporting the robustness of region-level evidence under reasonable threshold choices.

\begin{figure*}[!t]
    \centering
    \begin{overpic}[width=\textwidth]{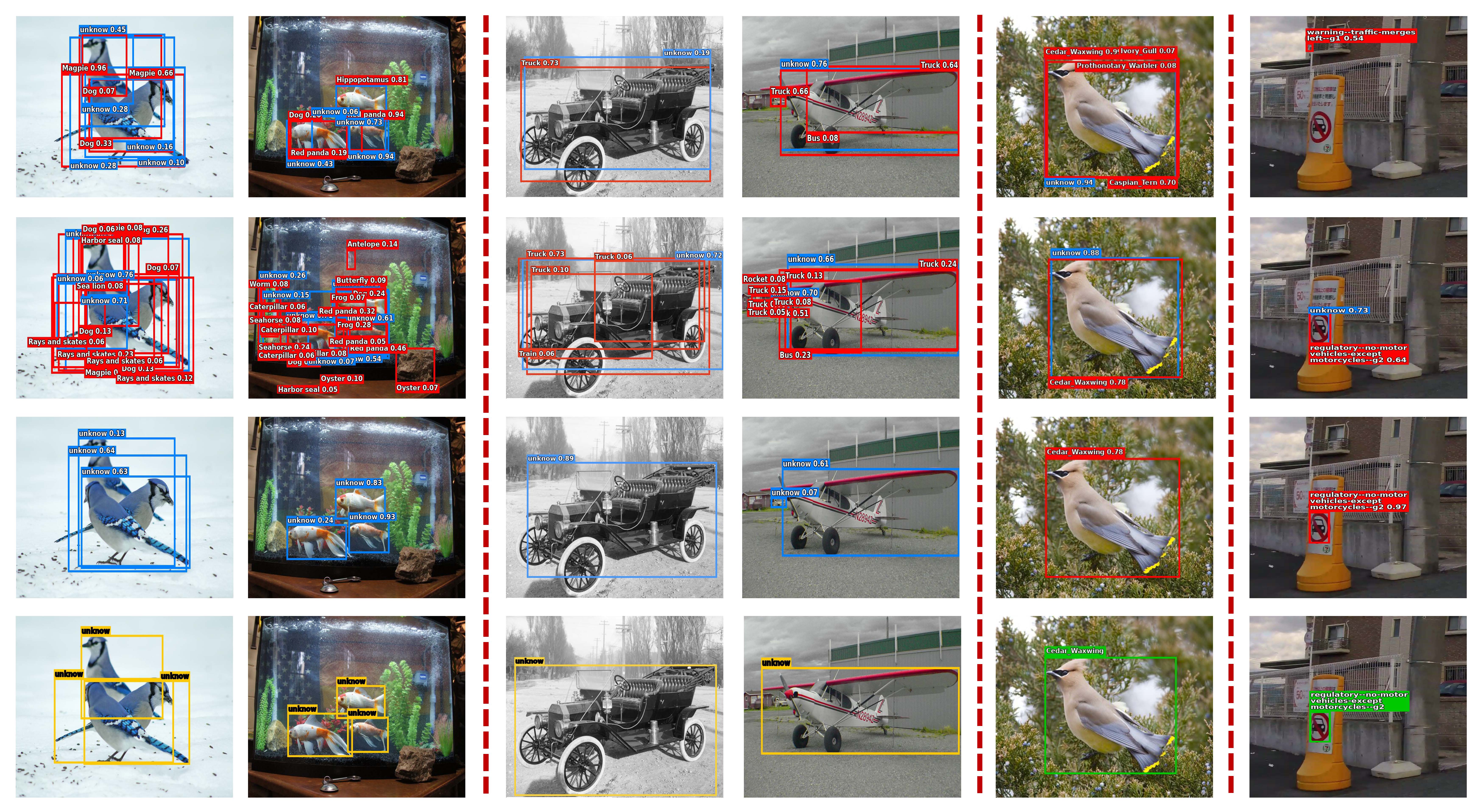}
    \put(10, -1){\scriptsize (a) OpenImages-Animal}
    \put(45, -1){\scriptsize (b) OpenImages-Vehicle}
    \put(72, -1){\scriptsize (c) CUB}
    \put(88, -1){\scriptsize (d) MTSD}
    \put(-1, 37){{\scriptsize \rotatebox{90}{Faster R-CNN~\citep{ren2015faster}}}}
    \put(-1, 26){{\scriptsize \rotatebox{90}{OpenDet~\citep{Han_2022_CVPR}}}}
    \put(-1, 15){{\scriptsize \rotatebox{90}{OCSI-Net}}}
     \put(-1, 5){{\scriptsize \rotatebox{90}{GT}}}
    \end{overpic}
    \vspace{0.5mm}
    \caption{Qualitative results on four datasets. The competitors produce conflicting known (red) and unknown (blue) predictions within the same region, while OCSI-Net enforces region-consistent semantics and yields coherent detections.}
    \label{fig:samples2}
\end{figure*}

\subsection{Qualitative Analysis}
Fig.~\ref{fig:samples2} presents the qualitative results. Across OpenImages-Animal/Vehicle, CUB200, and MTSD, both Faster R-CNN and OpenDet produce fragmented outputs in compact neighborhoods: multiple boxes covering the same underlying object are assigned contradictory semantics, where confident known-category predictions are interleaved with unknown responses. Such conflicts are especially pronounced in fine-grained or cluttered scenes, where slight spatial shifts among overlapping RoIs can flip the open-set decision and yield unstable, hard-to-interpret detections.

By contrast, our OCSI-Net exhibits substantially improved region-level coherence across all four datasets. As shown in the third row of Fig.~\ref{fig:samples2}, predictions within a compact spatial region are markedly less fragmented: overlapping detections largely agree on known/unknown semantics, and contradictory responses are effectively suppressed. This behavior confirms that OCSC shifts open-set reasoning from independent proposal-level decisions to region-consistent inference over clustered RoIs, preventing semantically incompatible predictions from coexisting within the same region. As a result, the final outputs are more coherent and better aligned with the correct semantic interpretation, especially in cases where baseline detectors produce conflicting predictions around a single object.

\begin{table*}[!t]
\caption{Analysis of different OCSC region encoders on CUB200. We report $\mathrm{AP}_{\mathrm{known}}$ and $\mathrm{AP}_{\mathrm{unk}}$ on each official split. The default frozen DINOv3 setting is highlighted.}
\label{tab:lim_encoder}
\centering
\tiny
\setlength{\tabcolsep}{2.3pt}
\renewcommand{\arraystretch}{1.05}
\resizebox{\textwidth}{!}{%
\begin{tabular}{>{\kern-\tabcolsep}lcccccccc<{\kern-\tabcolsep}}
\toprule
& \multicolumn{2}{c}{\textbf{Split 1}} & \multicolumn{2}{c}{\textbf{Split 2}} & \multicolumn{2}{c}{\textbf{Split 3}} & \multicolumn{2}{c}{\textbf{Split 4}} \\
\cmidrule(lr){2-3}\cmidrule(lr){4-5}\cmidrule(lr){6-7}\cmidrule(lr){8-9}
Method / OCSC region encoder
& $\mathrm{AP}_{\mathrm{known}}$ & $\mathrm{AP}_{\mathrm{unk}}$
& $\mathrm{AP}_{\mathrm{known}}$ & $\mathrm{AP}_{\mathrm{unk}}$
& $\mathrm{AP}_{\mathrm{known}}$ & $\mathrm{AP}_{\mathrm{unk}}$
& $\mathrm{AP}_{\mathrm{known}}$ & $\mathrm{AP}_{\mathrm{unk}}$ \\
\midrule
\rowcolor{blue!10} DINOv3~\cite{dino_v3_2025} frozen (Ours) & 63.8 & 35.1 & 62.9 & 45.2 & 64.3 & 41.2 & \textbf{62.6} & 40.4 \\
DINOv3~\cite{dino_v3_2025} (unfrozen) & \textbf{64.0} & \textbf{36.5} & \textbf{63.4} & \textbf{45.6} & \textbf{64.5} & \textbf{41.2} & 62.5 & \textbf{41.0} \\
CLIP-ViT-B/32~\citep{radford21clip} (frozen) & 63.7 & 31.7 & 62.6 & 40.5 & 64.1 & 37.1 & 62.2 & 39.3 \\
CLIP-ViT-B/32~\citep{radford21clip} (unfrozen) & 63.7 & 32.6 & 62.6 & 41.8 & 64.2 & 37.8 & 62.3 & 40.0 \\
DeiT-Tiny~\citep{touvron21deit} (frozen) & 63.7 & 32.1 & 62.5 & 41.5 & 64.1 & 35.7 & 62.1 & 38.5 \\
DeiT-Tiny~\citep{touvron21deit} (unfrozen) & 63.8 & 32.7 & 62.7 & 41.8 & 64.3 & 37.9 & 62.3 & 39.3 \\
ResNet-50~\citep{he16resnet} (frozen) & 63.7 & 31.2 & 62.5 & 40.4 & 64.1 & 34.6 & 62.1 & 38.0 \\
ResNet-50~\citep{he16resnet} (unfrozen) & 63.7 & 31.0 & 62.5 & 40.6 & 64.2 & 34.8 & 62.2 & 37.5 \\
\bottomrule
\end{tabular}}
\end{table*}


\section{Limitations and Analysis}
\label{sec:limitations}

\begin{figure*}[!t]
  \centering
     \begin{overpic}[width=\textwidth]{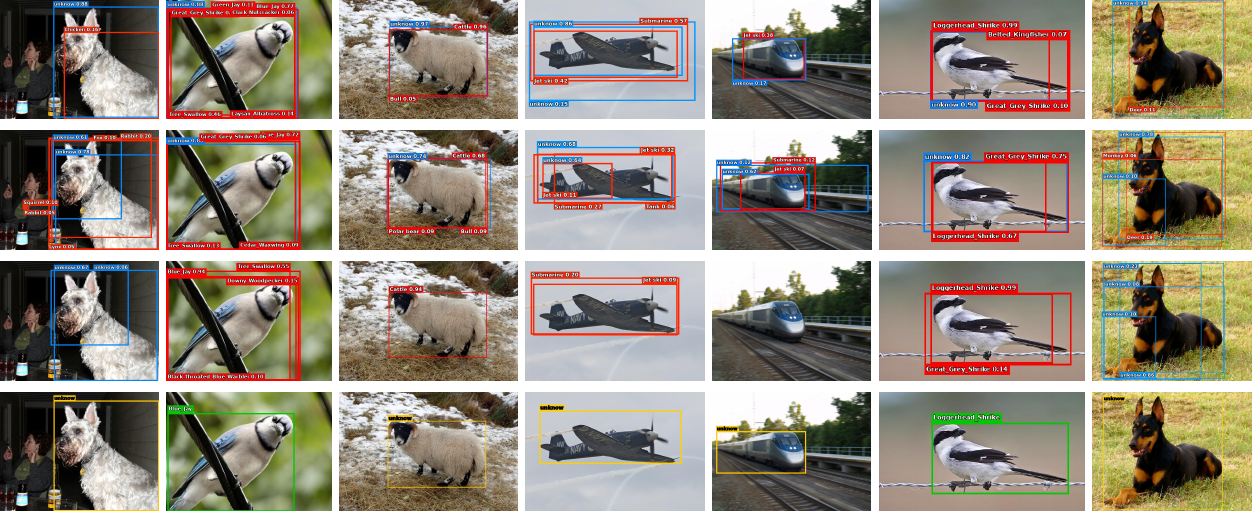}
    \put(-2, 31){{\scriptsize \rotatebox{90}{Faster R-CNN~\citep{ren2015faster}}}}
    \put(-2, 22){{\scriptsize \rotatebox{90}{OpenDet~\citep{Han_2022_CVPR}}}}
    \put(-2, 12){{\scriptsize \rotatebox{90}{OCSI-Net}}}
     \put(-2, 4){{\scriptsize \rotatebox{90}{GT}}}
     \end{overpic}
  \caption{Representative failure cases. These examples indicate that region-consistent refinement may be less effective when the region-level evidence is weak or unstable, or when crowded and heavily overlapping scenes require finer-grained instance-level discrimination.}
  \label{fig:failure_cases}
\end{figure*}

Despite the consistent improvements obtained by OCSI-Net, the proposed formulation introduces three practical limitations that deserve explicit discussion. First, OCSC relies on an auxiliary region encoder to extract semantic evidence from clustered proposals. Although this design is effective for resolving proposal-level inconsistencies, the use of a large pretrained visual encoder raises a natural question about the strictness of the open-set assumption. Second, the auxiliary evidence branch incurs additional inference cost. Third, the region-level semantic decision may still fail when the cluster evidence is weak, mixed, or insufficiently instance-aware. The following analysis clarifies these aspects and delineates the practical trade-offs of the proposed framework.


\subsection{Region Encoder and Open-Set Assumptions}
\label{subsec:limitation_encoder}

OCSC uses a frozen DINOv3~\citep{dino_v3_2025} encoder as the default region-evidence extractor. This choice provides strong region representations and improves the reliability of cluster-level knownness estimation. Nevertheless, when a foundation encoder is pretrained on large-scale external image corpora, the benchmark-level definition of ``unknown'' categories no longer guarantees that the visual appearances of these categories were absent during pretraining. This is an inherent limitation of using external representation priors in open-set detection: even if no label information from unknown categories is used during detector training, the frozen encoder may still contain general visual features learned from images that overlap semantically with the evaluation categories.

\textit{It is important to emphasize that this limitation is fundamentally different from direct label leakage.} In our framework, DINOv3~\cite{dino_v3_2025} is used only as a frozen feature extractor in the auxiliary OCSC branch. The base detector remains trained under the standard OSOD protocol~\citep{hosoya25rethinkingosod}, and the OCSC head is supervised only with known-region clusters. Therefore, the method does not access unknown-category labels or use unknown objects as supervised positives. The possible leakage risk lies instead in the representation prior inherited from external pretraining. Under this interpretation, the frozen encoder should be viewed as a source of generic visual evidence rather than as a category-level classifier over the unknown set.

To examine whether the proposed region-level semantic decision is tied to DINOv3 specifically, we compare multiple region encoders under frozen and unfrozen settings on the CUB200~\citep{wah2011caltech} splits. As shown in Table~\ref{tab:lim_encoder}, DINOv3 provides the strongest overall unknown-category performance, confirming that a stronger encoder improves the upper bound of region evidence. However, the same OCSC formulation remains functional with alternative encoders, including CLIP-ViT-B/32~\citep{radford21clip}, DeiT-Tiny~\citep{touvron21deit}, and ResNet-50~\citep{he16resnet}. Across these variants, $\mathrm{AP}_{\mathrm{known}}$ remains stable, while $\mathrm{AP}_{\mathrm{unk}}$ decreases as the region representation becomes weaker. This pattern suggests that the main mechanism of OCSC is not exclusive to a particular pretrained model, although its absolute performance is affected by the quality of the auxiliary visual representation.


These results indicate that the possible unknown-category exposure of DINOv3 should be interpreted as a limitation of external visual pretraining rather than as the sole reason for the effectiveness of OCSI-Net. A stricter evaluation of open-set purity would require auditing or controlling the pretraining corpus with respect to the benchmark unknown categories, or replacing the auxiliary encoder with a representation trained under category-disjoint constraints.

\subsection{Qualitative Failure Analysis}
\label{subsec:limitation_qualitative}



Fig.~\ref{fig:failure_cases} presents representative failure cases. When the region evidence incorrectly favors a known hypothesis for a truly unknown object, the imposed consistency constraint may suppress unknown responses within the same object region and retain an incorrect known prediction. In more difficult cases, weak or unstable evidence may also lead to overly aggressive suppression, resulting in missed detections.

These cases indicate three main limitations. First, in crowded or heavily overlapping scenes, proposals from different objects may be grouped into the same cluster, causing mixed evidence and propagating an incorrect semantic decision. Second, under poor localization, the envelope crop may include irrelevant context, weakening the region-level evidence used for refinement. Third, the current refinement primarily reconciles known--unknown conflicts; it does not explicitly resolve conflicts among known classes or perform instance-level de-duplication when multiple unknown hypotheses overlap. 


\section{Conclusion}
\label{sec:conclusion}

In this paper, we have proposed OCSI-Net, a region-consistent open-set detection framework that elevates open-set reasoning from individual proposals to region-level inference. Through object-centric semantic consistency inference and geometry-aware boundary uncertainty learning, OCSI-Net enforces semantic consistency and stabilizes open-set training. Experiments across multiple benchmarks demonstrate that OCSI-Net achieves state-of-the-art performance with a better balance between known and unknown detection.

\section*{Acknowledgment}
This work was supported in part by the Key R\&D  Program of Zhejiang Province 2026C01016, the National Natural Science Foundation of China under Grant 62576313, CASIC Guangxin Intelligent Technology, and Key R\&D  Program of Ningbo under Grant 2025Z128.

\bibliographystyle{IEEEtranN}
\bibliography{main}

\end{document}